\documentclass[twocolumn, switch]{article} % Method A for two-column formatting

\usepackage{preprint}

\usepackage{amsmath, amsthm, amssymb, amsfonts}
\usepackage{bm}
\usepackage{amsmath}
\usepackage{graphicx}

\usepackage{setspace}
\usepackage{amssymb}
\usepackage{booktabs}
\usepackage{array}
\usepackage{color}
\usepackage{multirow}
\usepackage{multicol}
\usepackage{url}
\usepackage[page]{appendix}

\usepackage[numbers,square]{natbib}
\usepackage[utf8]{inputenc}	% allow utf-8 input
\usepackage[T1]{fontenc}	% use 8-bit T1 fonts
\usepackage{xcolor}		% colors for hyperlinks
\usepackage[colorlinks = true,
            linkcolor = blue,
            urlcolor  = blue,
            citecolor = blue,
            anchorcolor = black]{hyperref}	% Color links to references, figures, etc.
\usepackage{booktabs} 		% professional-quality tables
\usepackage{nicefrac}		% compact symbols for 1/2, etc.
\usepackage{microtype}		% microtypography
\usepackage{lineno}		% Line numbers
\usepackage{float}			% Allows for figures within multicol
\usepackage{newfloat}
\DeclareFloatingEnvironment[name={Supplementary Figure}]{suppfigure}
\usepackage{sidecap}
\sidecaptionvpos{figure}{c}

\usepackage{titlesec}
\titlespacing\section{0pt}{12pt plus 3pt minus 3pt}{1pt plus 1pt minus 1pt}
\titlespacing\subsection{0pt}{10pt plus 3pt minus 3pt}{1pt plus 1pt minus 1pt}
\titlespacing\subsubsection{0pt}{8pt plus 3pt minus 3pt}{1pt plus 1pt minus 1pt}

\usepackage{graphics}
\usepackage{color}
\usepackage{hhline}
\usepackage{lipsum}

\usepackage{hyperref}
\usepackage{bm}
\usepackage[flushleft]{threeparttable}
\usepackage{bbding}
\usepackage{float}
\usepackage{subfig}
\usepackage{multirow}
\usepackage[inline]{enumitem}
\usepackage[ruled]{algorithm2e}

\title{Diffusion-based Perceptual Neural Video Compression with Temporal Diffusion Information Reuse}

\usepackage{authblk}

\author[ ]{Wenzhuo Ma}
\author[*]{Zhenzhong Chen}
\affil[ ]{School of Remote Sensing and Information Engineering, Wuhan University}

\begin{document}

\twocolumn[ % Method A for two-column formatting
  \begin{@twocolumnfalse} % Method A for two-column formatting
  
\maketitle

\begin{abstract}

Recently, foundational diffusion models have attracted considerable attention in image compression tasks, whereas their application to video compression remains largely unexplored. In this article, we introduce DiffVC, a diffusion-based perceptual neural video compression framework that effectively integrates foundational diffusion model with the video conditional coding paradigm. This framework uses temporal context from previously decoded frame and the reconstructed latent representation of the current frame to guide the diffusion model in generating high-quality results. To accelerate the iterative inference process of diffusion model, we propose the Temporal Diffusion Information Reuse (TDIR) strategy, which significantly enhances inference efficiency with minimal performance loss by reusing the diffusion information from previous frames. Additionally, to address the challenges posed by distortion differences across various bitrates, we propose the Quantization Parameter-based Prompting (QPP) mechanism, which utilizes quantization parameters as prompts fed into the foundational diffusion model to explicitly modulate intermediate features, thereby enabling a robust variable bitrate diffusion-based neural compression framework. Experimental results demonstrate that our proposed solution delivers excellent performance in both perception metrics and visual quality.

\end{abstract}

\vspace{0.4cm}

  \end{@twocolumnfalse} % Method A for two-column formatting
] % Method A for two-column formatting

\newcommand\blfootnote[1]{%
\begingroup
\renewcommand\thefootnote{}\footnote{#1}%
\addtocounter{footnote}{-1}%
\endgroup
}

\section{INTRODUCTION}

{\blfootnote{Corresponding author: Zhenzhong Chen, E-mail:zzchen@ieee.org}}
  
With the rise of the digital age, multimedia content, especially video, has become a major component of internet traffic. Consequently, for more efficient transmission and storage, video compression technology has become a research focus. Over the past few decades, traditional coding standards such as AVC \cite{DBLP:journals/tcsv/WiegandSBL03}, HEVC \cite{DBLP:journals/tcsv/SullivanOHW12} and VVC \cite{DBLP:journals/tcsv/BrossWYLCSO21} have been developed and widely adopted. Recently, learning-based neural video compression (NVC) has shown remarkable performance, with methods like DCVC-DC \cite{DBLP:conf/cvpr/LiLL23} and DCVC-FM \cite{DBLP:conf/cvpr/LiLL24} outperforming the best traditional codec ECM. Most NVC methods are optimized for the trade-off between bitrate and pixel-level distortion (such as mean squared error). However, Blau et al. \cite{DBLP:conf/icml/BlauM19} demonstrate that pixel-level distortion does not correspond to human visual perception. In other words, reconstruction results with lower pixel-level distortion can still appear blurry and unrealistic, particularly at lower bitrates.

Perceptual video compression seeks to optimize for rate-perception trade-off, producing more realistic outcomes. Previously, perceptual video compression approaches fall into two main categories: one \cite{DBLP:journals/pr/XuLYLZ24} uses perceptual loss functions (e.g. LPIPS \cite{DBLP:conf/cvpr/ZhangIESW18}) during training to enhance perceptual quality while the others \cite{DBLP:conf/eccv/MentzerABMJT22,DBLP:conf/vcip/ZhangMHBWY21,DBLP:conf/ijcai/YangTG22,DBLP:conf/mm/LiSWH23,DBLP:journals/esticas/DuLL24} rely on GAN-based frameworks to leverage the generative capabilities of GANs \cite{DBLP:conf/nips/GoodfellowPMXWOCB14} for detail-rich reconstructions. Recently, the development of foundational diffusion models (e.g. Stable Diffusion \cite{DBLP:conf/cvpr/RombachBLEO22}) has opened new possibilities for perception-oriented compression. These foundational diffusion models, trained on thousands of high-quality image-text pairs, can generate high-quality, clear images. Some methods \cite{DBLP:journals/corr/abs-2211-07793,DBLP:conf/nips/YangM23,DBLP:journals/corr/abs-2307-01944,DBLP:conf/iclr/CareilMVL24,DBLP:journals/corr/abs-2404-08580,DBLP:journals/corr/abs-2410-02640} have applied foundational diffusion models to image compression tasks, leveraging their powerful generative capabilities to achieve significant improvements in visual perception of reconstructed results. This naturally inspires the idea of incorporating foundational diffusion models into video compression frameworks to produce reconstructed videos with high perceptual quality.

Incorporating foundational diffusion models into video compression tasks presents three main challenges: 
\begin{enumerate}[label = (\arabic*)]
	\item Develop a framework for effectively integrating foundational diffusion models into the state-of-the-art conditional coding paradigm for video compression.
	\item The iterative nature of diffusion model inference introduces high latency. If each frame requires numerous diffusion steps, the resulting delay is unacceptable for video applications. Therefore, efficient inference strategies are crucial for using diffusion models in video compression.
	\item Variable bitrate is a key feature of video codecs, yet distortion levels of latent representations vary across bitrates, posing a challenge for diffusion models. Therefore, enabling diffusion models to perceive and adapt to distortion variations of latent representations is one of the key challenges in achieving robust variable bitrate functionality for diffusion-based video compression methods.
\end{enumerate}

To effectively integrate foundational diffusion models into a conditional coding framework, we propose a perceptual neural video compression framework based on diffusion model, named DiffVC, which uses temporal context extracted from the previous decoded frame and reconstructed latent representations decoded from the bitstream of the current frame as conditions to guide the diffusion model in generating high-quality results. 
To accelerate inference, we propose an efficient inference strategy based on temporal diffusion information reuse. Given the substantial correlation between consecutive video frames, the current frame can partially reuse diffusion information from the previous frames, thereby significantly expediting the inference process. Specifically, the diffusion process for each P-frame is divided into two stages: the first stage reuses diffusion information from the previous frames for rapid processing, while the second stage employs conventional diffusion steps to reconstruct high-quality details. Experimental results show that this temporal diffusion information reuse strategy reduces inference time by 47\%, with only a 1.96\% perceptual performance loss. This significant improvement in inference speed is achieved with minimal trade-offs.
To address distortion differences across various bitrates, we employ a simple but effective quantization parameter-based prompting mechanism to modulate the diffusion model. Specifically, we feed the quantization parameters generated by the compression model as prompts to the foundational diffusion model. By leveraging the diffusion model's ability to interpret semantic information and applying targeted fine-tuning, this mechanism enables the diffusion-based video compression network to adapt effectively to distortions across different bitrates.
Extensive experiments demonstrate that DiffVC excels in perception metrics and visual quality. Notably, the proposed DiffVC achieves state-of-the-art performance across all test datasets for the DISTS \cite{DBLP:journals/pami/DingMWS22} metric. Benefiting from the temporal diffusion information reuse strategy and the quantization parameter-based prompting mechanism, DiffVC achieves efficient inference and robust variable bitrate functionality with a single model.

Our contributions are summarized as follows: 
\begin{itemize}
	\item We propose DiffVC, a diffusion-based perceptual neural video compression framework. It integrates the foundational diffusion model with the conditional coding paradigm by using temporal context from the previously decoded frame and reconstructed latent representations from the current frame's bitstream as conditions to generate high-quality results.
	\item We introduce an efficient inference strategy based on temporal diffusion information reuse, which achieves a significant improvement in inference speed with minimal perceptual performance degradation.
	\item We introduce a quantization parameter-based prompting mechanism that explicitly modulates the diffusion model using quantization parameters as prompts, enabling DiffVC to roboustly support variable bitrates.
	\item Experimental results on several datasets demonstrate that DiffVC delivers remarkable performance in perception metrics and visual quality, particularly achieving optimal performance across all datasets for the DISTS metric.
\end{itemize}

\section{Related Work} \label{sec:related_work}
\subsection{Neural Video Compression}
With the impressive performance of deep neural networks in image compression tasks \cite{DBLP:conf/iclr/BalleLS17,DBLP:conf/iclr/BalleMSHJ18,DBLP:conf/nips/MinnenBT18,DBLP:conf/cvpr/ChengSTK20,DBLP:conf/cvpr/HeZSWQ21,DBLP:conf/cvpr/HeYPMQW22,DBLP:conf/cvpr/LiuSK23,DBLP:journals/tomccap/ZhaoLLXL24} , learning-based neural video compression has been widely researched. Existing neural video compression methods can be roughly categorized into three types: residual coding-based, 3D autoencoder-based, and conditional coding-based. Residual coding-based methods \cite{DBLP:conf/cvpr/LuO0ZCG19,DBLP:conf/cvpr/HuL021,DBLP:journals/pami/LuZO0G021,DBLP:conf/ijcai/YangTG22,DBLP:journals/tmm/WangCC24,DBLP:journals/tomccap/YangYXZW24} first generate a predicted frame based on the previously decoded frame, and then encode the residue between the current frame and the predicted frame. Lu et al. \cite{DBLP:conf/cvpr/LuO0ZCG19} proposed the first neural video compression method, which is residual coding-based. This approach uses traditional codecs as a template, replacing all of their modules with neural networks and jointly training them in an end-to-end manner. However, the residual coding-based approach, which merely reduces inter-frame redundancy through simple subtraction operation, is not thorough and is suboptimal. The 3D autoencoder-based methods \cite{DBLP:conf/iccv/HabibianRTC19,DBLP:conf/eccv/SunTLYYL20,DBLP:conf/cvpr/VeerabadranPHC20} extend the autoencoder in image compression tasks by treating video as multiple images with a temporal dimension, but it introduces significant encoding delays and substantially increases memory costs. The conditional coding-based methods \cite{DBLP:conf/mm/LiSWH23,DBLP:journals/tip/WangC24,DBLP:journals/pr/XuLYLZ24,DBLP:conf/nips/LiLL21,DBLP:conf/mm/Li0022,DBLP:journals/tmm/ShengLLLLL23,DBLP:conf/cvpr/LiLL23,DBLP:conf/cvpr/LiLL24} extract contextual information from the previously decoded frame in the feature domain and uses it as a condition to assist the encoder, decoder, and entropy model during encoding and decoding. Conditional coding-based methods achieve superior compression performance by avoiding constraints on the context to the pixel domain, allowing them to learn richer context and eliminate inter-frame redundancy. The well-known DCVC series \cite{DBLP:conf/nips/LiLL21,DBLP:conf/mm/Li0022,DBLP:journals/tmm/ShengLLLLL23,DBLP:conf/cvpr/LiLL23,DBLP:conf/cvpr/LiLL24} adopts the conditional coding-based paradigm, with DCVC-DC \cite{DBLP:conf/cvpr/LiLL23} and DCVC-FM \cite{DBLP:conf/cvpr/LiLL24} even surpassing the best traditional codec ECM.

\subsection{Perceptual Video Compression}
Although neural video compression methods have shown excellent performance in pixel-level distortion metrics, the work by Blau et al. \cite{DBLP:conf/icml/BlauM19} demonstrates the existence of a "rate-distortion-perception" trade-off, suggesting that better perceptual quality at a fixed rate often corresponds to greater distortion. As a result, perception-oriented neural video compression has attached significant attention. The work of \cite{DBLP:journals/pr/XuLYLZ24} incorporated perceptual loss terms, such as Learned Perceptual Image Patch Similarity (LPIPS) \cite{DBLP:conf/cvpr/ZhangIESW18}, into the rate-distortion loss functions, optimizing for visual quality. Additionally, some methods leverage the powerful generative capabilities of Generative Adversarial Network (GAN) \cite{DBLP:conf/nips/GoodfellowPMXWOCB14} to develop neural video codecs that can produce realistic, high-perceptual-quality reconstruction results. Mentzer et al. \cite{DBLP:conf/eccv/MentzerABMJT22} pioneered the introduction of GANs into neural video compression, which treats the compression network as a generator. Through the adversarial process with a discriminator, the network learns to reconstruct videos with rich details. Zhang et al. \cite{DBLP:conf/vcip/ZhangMHBWY21} introduced a discriminator and hybrid loss function based on DVC \cite{DBLP:conf/cvpr/LuO0ZCG19} to help the network trade off rate, distortion and perception. Yang et al. \cite{DBLP:conf/ijcai/YangTG22} introduced a recurrent conditional GAN, which consists of a recurrent generator and a recurrent discriminator conditioned on the latent representations generated during compression, resulting in remarkable perceptual quality outcomes. To address the issue of poor reconstruction quality in newly emerged areas and the presence of checkerboard artifacts in GAN-based methods, Li et al. \cite{DBLP:conf/mm/LiSWH23} designed a confidence-based feature reconstruction method, combined with a periodic compensation loss function, which further improves the visual quality of the reconstructed video. Du et al. \cite{DBLP:journals/esticas/DuLL24} proposed a contextual generative video compression method with transformers, named CGVC-T, which employs GAN to enhance perceptual quality and utilizes contextual coding to improve compression efficiency.

\subsection{Diffusion-based Compression}
Recently, benefiting from the application of pre-trained foundational diffusion models (such as Stable Diffusion \cite{DBLP:conf/cvpr/RombachBLEO22}) trained on thousands of high-quality image-text pairs, diffusion-based image compression methods \cite{DBLP:journals/corr/abs-2211-07793,DBLP:conf/nips/YangM23,DBLP:journals/corr/abs-2307-01944,DBLP:conf/iclr/CareilMVL24,DBLP:journals/corr/abs-2404-08580,DBLP:journals/corr/abs-2410-02640} have surpassed GAN-based methods in perception metrics. Lei et al. \cite{DBLP:journals/corr/abs-2307-01944} proposed to only transmit the sketch and text description of the image, then use both as conditions to guide the diffusion model to generate the reconstructed image during decoding. The work by Careil et al. \cite{DBLP:conf/iclr/CareilMVL24} guided the diffusion decoding process using vector-quantized latent representations and image descriptions. Relic et al. \cite{DBLP:journals/corr/abs-2404-08580} treated the removal of quantization noise as a denoising task and employed a parameter estimation module to learn adaptive diffusion steps, which achieved high-quality results with only 2\% to 7\% of the full diffusion process. Li et al. \cite{DBLP:journals/corr/abs-2410-02640} used compressed latent features with added noise instead of pure noise as the starting point, significantly reducing the number of diffusion steps required for reconstruction and introduced a novel relay residual diffusion process to further enhance reconstruction quality.

Despite the significant success of diffusion models in image compression, diffusion-based video compression methods have been rarely studied. As discussed earlier, incorporating diffusion models into neural video compression presents three key challenges. First, there is the need to effectively integrate foundational diffusion models into the video compression framework to enhance the perceptual quality of reconstruction without disrupting existing coding paradigms, such as conditional coding. Li et al. \cite{DBLP:journals/corr/abs-2402-08934} proposed a hybrid approach that combines image compression and diffusion models for video compression. However, this fragmented strategy disrupted video coding paradigms, resulting in suboptimal reconstruction performance. Second, the slow inference speed of diffusion-based video compression methods needs to be addressed. For example, the work by Liu et al. \cite{liu2024i2vcunifiedframeworkintra} proposed a novel diffusion-based video compression framework that integrates different video compression modes (such as AI, LDP, LDB, and RA) into a unified system. However, it faces significant challenges in terms of inference latency. Finally, the varying distortion levels of latent representations across different bitrates pose an additional challenge: enabling a single diffusion model to support inference across multiple bitrates. The proposed DiffVC effectively addresses these three issues, realizing a diffusion-based perceptual neural video compression framework that supports efficient inference and roboust variable bitrate.

\section{Methodology} \label{sec:method}
\begin{table}[th]
	\caption{  A list of notations mainly used in this paper.}
	\centering
	{
		\resizebox{1.0\linewidth}{!}{
			\begin{tabular}{|l|l|}
				\hline
				Symbol & Definition \\
				\hline
				\hline
				$t$                		& Video frame index. $t \in [0,T)$ \\ \hline
				$n$                		& Diffusion timestep. $n \in [0,N]$ \\ \hline
				$DS$               		& Total number of diffusion steps. \\ \hline
				$D$                		& The number of independent diffusion steps. \\ \hline
				$ds$                	& Diffusion step index. $ds \in [0,DS]$ \\ \hline
				$x_t$  			   		& Original input frame. \\ \hline
				$\hat{x}_t$        		& Reconstructed frame. \\ \hline
				$m_t$              		& Estimated motion vector. \\ \hline
				$\hat{m}_t$        		& Reconstructed motion vector. \\ \hline
				$y_t$              		& The latent representation output by $\mathcal{E}$. \\ \hline
				$\hat{y}_t$        		& Reconstructed latent representation. \\ \hline
				$\bar{C}_t$        		& Multi-scale temporal contexts mined from previous decoded frame. \\ \hline
				$f_t$              		& The feature of reconstructed latent representation $\hat{y}_t$. \\ \hline
				$z_t^n$            		& Noisy latent representation of timestep $n$ in the diffusion process. \\ \hline
				$z_t^0$            		& Denoised latent representation of timestep $0$ in the diffusion process. \\ \hline
				$\ddot{y}_t^n$     		& The predicted noise-free latent of timestep $n$. \\ \hline
				$\epsilon^n$       		& Standard Gaussian noise added at timestep $n$. \\ \hline
				$\epsilon_{\theta}^n$   & The noise predicted by U-Net at timestep $n$. \\ \hline
				$q_{enc}$               & The quantization parameter used in Contextual Encoder. \\ \hline
				$q_{dec}$               & The quantization parameter used in Contextual Decoder. \\ \hline
				$ratio$            		& The ratio of quantized parameters, that is, QP-based prompt. \\ \hline
				$\mathcal{E}$      		& The encoder od pre-trained Stable Diffusion V2.1. \\ \hline
				$\mathcal{D}$      		& The decoder od pre-trained Stable Diffusion V2.1. \\ \hline
		\end{tabular}}
	}
	\label{tab:notation}
\end{table}
\begin{figure*}[!t]
	\centering
	\includegraphics[width=1.0\textwidth]{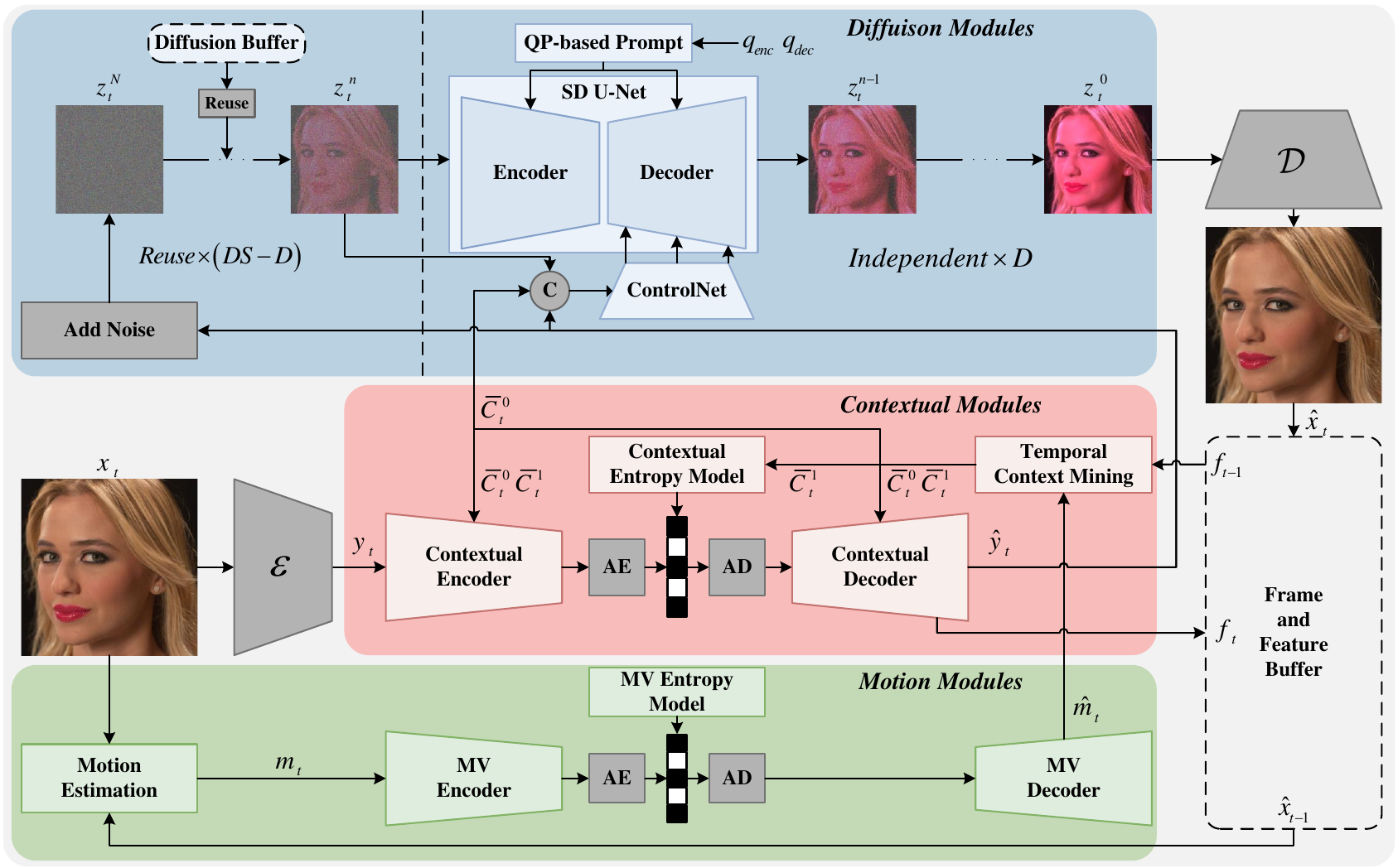}
	\caption{The framework of DiffVC. DiffVC consists of three main components: Motion Modules, Contextual Modules, and Diffusion Modules. The Motion Modules (green) manage motion vector estimation and compression. The Contextual Modules (red) focus on extracting temporal context and compressing conditional residues. Finally, the Diffusion Modules (blue) apply multiple diffusion steps to generate high perceptual-quality reconstructions. The components $\mathcal{E}$ and $\mathcal{D}$ represent the pre-trained autoencoder of Stable Diffusion V2.1. The Frame and Feature Buffer stores the previous decoded frame and its latent representation feature, while the Diffusion Buffer stores diffusion information from the previous frame.}
	\label{fig:DiffVC}
\end{figure*}

\subsection{The Framework of DiffVC} \label{sec:framework}
To effectively incorporate the foundational diffusion model within a conditional coding framework, we propose a diffusion-based perceptual neural video compression framework, named DiffVC. DiffVC consists of three main components: Motion Modules, Contextual Modules, and Diffusion Modules. The Motion and Contextual Modules are adapted from DCVC-DC \cite{DBLP:conf/cvpr/LiLL23}. As shown in Table \ref{tab:notation}, which defines the primary symbols used in this paper, the original video frames and decoded frames are denoted as $\{x_t\}^T_{t=1}$ and $\{\hat{x}_t\}^T_{t=1}$, respectively. The overall model architecture is depicted in Fig. \ref{fig:DiffVC}, the details of DiffVC are as follows:

\textbf{Motion Modules}. The motion vectors between the previously decoded frame and the current frame are estimated, then encoded and decoded.
\begin{itemize}
	\item Motion Estimation: A motion estimation network (Spynet \cite{DBLP:conf/cvpr/RanjanB17}) estimates the optical flow $m_t$ between the previously decoded frame $\hat{x}_{t-1}$ and the current frame $x_t$.
	\item MV Encoder / Decoder: The motion vector $m_t$ is encoded into the bitstream by an autoencoder and decoded to  obtain the reconstructed motion vector $\hat{m}_t$.
	\item MV Entropy Model: For simplicity, the entropy model for the motion vector includes only a hyperprior model and quadtree partition-based entropy coding, without using the latent representation of the previous frame as a prior.
\end{itemize}

\textbf{Contextual Modules}. The encoder $\mathcal{E}$ of Stable Diffusion V2.1 is first used to transform the current frame $x_t$ into the latent representation $y_t$. Then $y_t$ is compressed and decompressed with the aid of temporal context mined from reconstructed optical flow $\hat{m}_t$ and the latent representation feature $f_{t-1}$ of previously decoded frame.
\begin{itemize}
	\item Temporal Context Mining: The reconstructed latent representation feature $f_{t-1}$ of the previously decoded frame is aligned to current frame with the reconstructed motion vector $\hat{m}_t$. A hierarchical approach is then performed to learn multi-scale temporal contexts $\bar{C}_t$. Specifically, two scales of temporal contexts ($\bar{C}_t^0$ and $\bar{C}_t^1$) are extracted in DiffVC, which are $\frac{1}{8}$ and $\frac{1}{16}$ of the original resolution. It is worth noting that group-based offset diversity in DCVC-DC is not used in DiffVC considering the model complexity.
	\item Contextual Encoder / Decoder: With the assistance of the temporal contexts $\bar{C}_t^0$ and $\bar{C}_t^1$, the current frame’s latent representation $y_t$ is encoded into the bitstream and reconstructed as $\hat{y}_t$. Notably, since the $y_t$ has already been downsampled by a factor of 8, the Contextual Encoder / Decoder only applies a $2\times$ downsampling, ensuring the representation to be written into bitstream is $\frac{1}{16}$ of the original resolution.
	\item Contextual Entropy Model: The entropy model in the contextual modules employs hyperprior model and quadtree partition-based entropy coding, with the small-scale temporal context $\bar{C}_t^1$ and the latent representation decoded from the previous frame's bitstream as priors.
\end{itemize}

The Motion and Contextual Modules enable the conditional compression of video frames. The key challenge lies in integrating the foundational diffusion model into the conditional coding paradigm. Unlike diffusion-based image compression, where only the decoded latent representations serve as conditions to guide the diffusion model in generating the reconstructed result, diffusion-based video compression can exploit prior information from previously decoded frames due to the high redundancy between video frames. In the conditional coding paradigm, the Contextual Modules already extract rich, multi-level contextual information from the feature domain of previously decoded frames to assist in encoding and decoding the current frame. This "ready-made" contextual information can be reused to guide the diffusion model in reconstructing the current frame. By leveraging this approach, we seamlessly integrate the condition-guided foundational diffusion model into the conditional coding paradigm, enabling a conditional diffusion-based video compression framework. Specifically, the details of the Diffusion Modules in DiffVC are as follows:

\begin{figure*}[!t]
	\centering
	\includegraphics[width=1.0\textwidth]{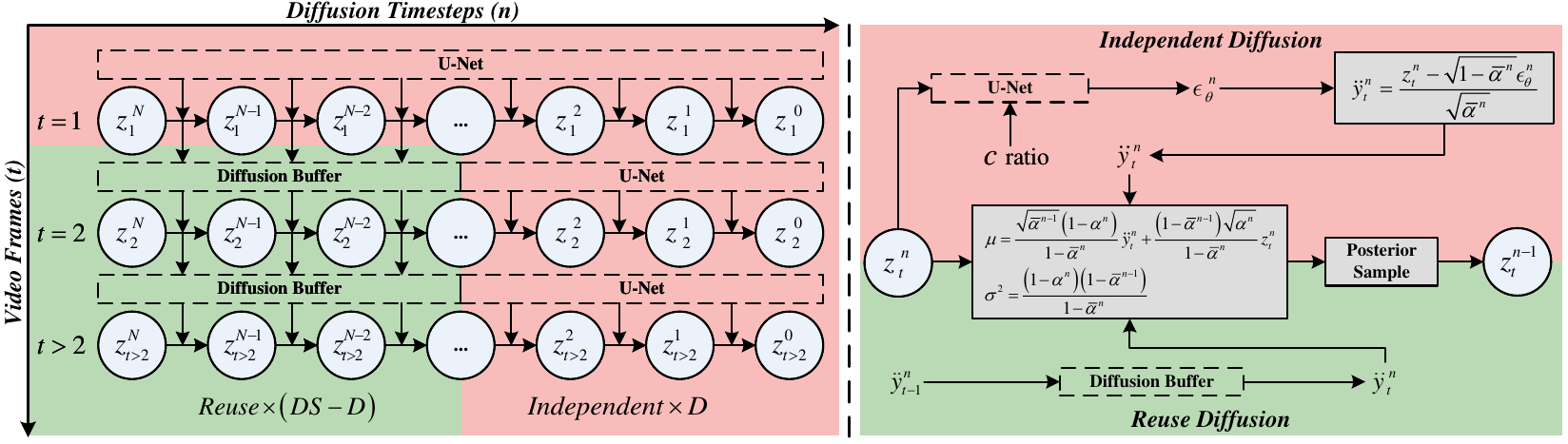}
	\caption{Temporal Diffusion Information Reuse Strategy. The left panel illustrates the inference process of TDIR, where the vertical axis represents video frames and the horizontal axis represents diffusion timesteps. The first P frame undergoes independent diffusion for $DS$ steps, while subsequent P frames reuse diffuse for $DS-D$ steps before undergoing independent diffusion for the remaining $D$ steps. The right panel provides details on the independent diffusion step (red) and reuse diffusion step (green).}
	\label{fig:TDIR}
\end{figure*}

\textbf{Diffusion Modules}: This modules begin by adding Gaussian noise over $N$ steps to the reconstructed latent representation $\hat{y}_t$, resulting in an initial diffusion state $z_t^N$. The diffusion process then applies the proposed temporal diffusion information reuse strategy to produce the denoised representation $z_t^0$, which will be  detailed in Section \ref{sec:TDIR}. Finally, the decoder $\mathcal{D}$ of Stable Diffusion V2.1 reconstructs the frame $\hat{x}_t$ with high perceptual quality.
\begin{itemize}
	\item Noise Estimation: The U-Net from Stable Diffusion V2.1 serves as the noise estimation network in DiffVC.
	\item Conditional Guidance: In DiffVC, the noised latent representation $z_t^n$ at the current timestep, the reconstructed latent representation $\hat{y}_t$ of the current frame, and the larger scale temporal context $\bar{C}_t^0$ mined from previously decoded frame are concatenated and used as conditional input to ControlNet \cite{DBLP:conf/iccv/ZhangRA23}, which guides the U-Net network in noise estimation.
	\item QP-based Prompt: To enable the U-Net to recognize distortion variations across bitrates, the quantization parameters ($q_{enc}$ and $q_{dec}$) used in the Contextual Encoder and Decoder are provided to the U-Net as prompts. This modulation is achieved through cross-attention layers, allowing the U-Net to adaptively adjust based on bitrate-dependent distortion levels. The details are presented in Section \ref{sec:QPP}.
\end{itemize}

Overall, the DiffVC framework tightly integrates the conditional coding framework with the foundational diffusion model by using contextual information as a bridge, leveraging the powerful prior knowledge of the diffusion model to reconstruct high-quality results with excellent perceptual fidelity. Notably, since the conditions guiding the diffusion model include both the current frame’s information ($\hat y_t$) and the information from previously decoded frames ($\bar{C}_t^0$), the diffusion model can more comprehensively account for video continuity during reconstruction, rather than merely recovering the current frame.

\subsection{Temporal Diffusion Information Reuse Strategy} \label{sec:TDIR}
Following DDPM \cite{DBLP:conf/nips/HoJA20}, DiffVC obtains the noisy latent $z_t^n$ by adding Gaussian noise with variance $\beta^n \in (0,1)$ to the latent $\hat{y}_t$ reconstructed by the Contextual Decoder:
\begin{equation}
	\label{eq:add_noise}
	z_t^n = \sqrt{\bar{\alpha}^n}\hat{y}_t + \sqrt{1-\bar{\alpha}^n}\epsilon^n,\ n = 0,1,2,...,N
\end{equation}
where $n \in [0,N]$ is diffusion timestep, $t \in [0,T)$ is video frame index, $\epsilon^n \in \mathcal{N}(0,\boldsymbol I)$ is standard Gaussian noise, $\beta^n$ is a fixed constant that increases as $n$ increases, $\alpha^n = 1 - \beta^n$ and $\bar{\alpha}^n=\prod_{i=1}^n \alpha^n$. When $n$ approaches $N$ (very large), the noisy latent $z_t^n$ is nearly a standard Gaussian distribution.

In conventional diffusion processes, the noisy latent representation undergoes multiple diffusion steps to produce the final denoised latent representation, which is then decoded to generate the reconstruction. However, applying this process to video compression tasks results in unacceptable inference latency due to the need for multiple iterations per frame. Fortunately, unlike standalone images, consecutive video frames share substantial temporal correlation, allowing the reuse of diffusion information across frames. Based on this idea, we propose a  Temporal Diffusion Information Reuse (TDIR) strategy, which significantly accelerates the inference process for diffusion models in video compression with only a minimal loss in perceptual quality.

\textbf{Overview.} TDIR utilizes two diffusion modes: Independent Diffusion and Reuse Diffusion (illustrated in Fig. \ref{fig:TDIR}, with the left panel showing the overall inference flow, the top-right corner depicting independent diffusion, and the bottom-right corner illustrating reuse diffusion). Assume each P frame requires a total of $DS$ diffusion steps, of which $D$ steps are independent diffusion, and the remaining $DS-D$ steps involve reuse diffusion. For the first P frame in each Group of Pictures, the process exclusively employs independent diffusion ($D=DS$). For subsequent P frames, the process starts with $DS - D$ reuse diffusion steps, followed by $D$ independent diffusion steps ($D=\frac{1}{2}DS$ in DiffVC).

\begin{figure*}[!t]
	\centering
	\includegraphics[width=1.0\textwidth]{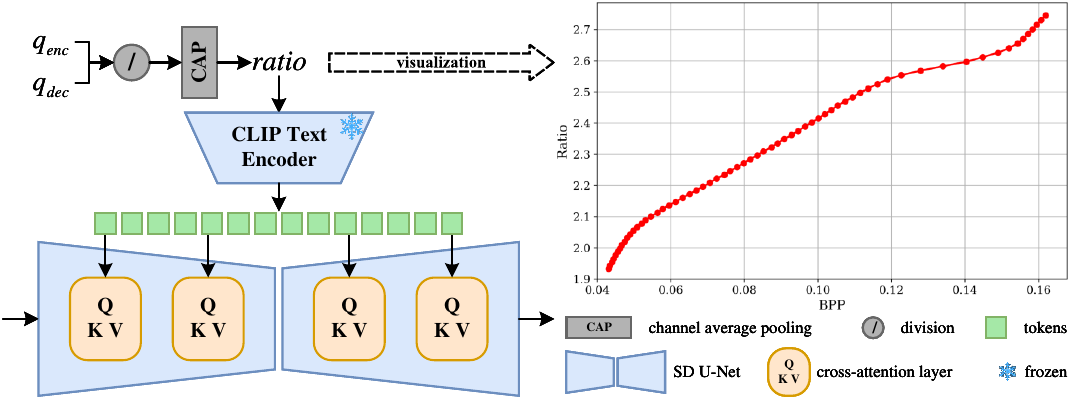}
	\caption{QP-based Prompting Mechanism. The ratio of the quantization parameters, $q_{enc}$ and $q_{dec}$, in the Contextual Encoder/Decoder, is averaged channel-wise and encoded into tokens using the pretrained CLIP Text Encoder. These tokens are then used to modulate the intermediate features of the U-Net via cross-attention layers. The visualization in the top-right corner illustrates the relationship between the quantization parameter ratio and the bitrate, with the test data being the \textit{BQMall} sequence from HEVC Class C.}
	\label{fig:QPP}
\end{figure*}

\textbf{Independent Diffusion.} At each independent diffusion step, the noisy latent $z_t^n$ is used as input. As described in Section \ref{sec:framework}, the noisy latent $z_t^n$, the reconstructed latent $\hat{y}_t$ and the large-scale temporal context $\bar{C}_t^0$ are concatenated together to form the condition $c$. Under the guidance of the specific condition $c$ and the QP-based prompt $ratio$ (detailed in Section \ref{sec:QPP}), the U-Net predicts the noise $\epsilon_{\theta}^n$ for the current timestep.
\begin{equation}
	\label{eq:predict_noise}
	\begin{aligned} 
		c &= concat(z_t^n, \hat y_t, \bar{C}_t^0) \\
		\epsilon_{\theta}^n &= Unet(z_t^n,c,ratio)
	\end{aligned}
\end{equation}
Subsequently, based on Eq. \ref{eq:add_noise}, we derive Eq. \ref{eq:predict_noise_free}, which enables the prediction of the noise-free latent $\ddot{y}_t^n$ for the current timestep.
\begin{equation}
	\label{eq:predict_noise_free}
	\ddot{y}_t^n = \frac{z_t^n - \sqrt{1-\bar{\alpha}^n}\epsilon_{\theta}^n}{\sqrt{\bar{\alpha}^n}}
\end{equation}
Then, the mean $\mu$ and variance $\sigma$ of the posterior distribution for the current timestep are calculated using $\ddot{y}_t^n$ and $z_t^n$:
\begin{equation}
	\label{eq:postrior_distribution}
	\begin{aligned} 
		\mu &= \frac{\sqrt{\bar{\alpha}^{n-1}}(1-\alpha^n)}{1-\bar{\alpha}^n}\ddot{y}_t^n + \frac{(1-\bar{\alpha}^{n-1})\sqrt{\alpha^n}}{1-\bar{\alpha}^n}z_t^n \\
		\sigma^2 &= \frac{(1-\alpha^n)(1-\bar{\alpha}^{n-1})}{1-\bar{\alpha}^n}
	\end{aligned}
\end{equation}
Finally, posterior sampling is conducted to obtain the latent $z_t^{n-1}$ for timestep $n-1$:
\begin{equation}
	\label{eq:postrior_sample}
	z_t^{n-1} \sim \mathcal{N}(\mu, \sigma^2 \boldsymbol I)
\end{equation}
where $\boldsymbol I$ is identity matrix. This completes a single independent diffusion step. However, since noise prediction relies on the highly complex and parameter-intensive U-Net network, the independent diffusion mode is relatively slow.

\textbf{Reuse Diffusion.} Owing to the temporal correlation between video frames, the predicted noise-free latents $\ddot{y}_t^n$ at corresponding timesteps in the diffusion processes of adjacent frames are highly similar (as shown in Fig. \ref{fig:cosine_similarity}). Leveraging this, the Reuse Diffusion mode accelerates the diffusion process for the current frame by reusing the predicted noise-free latents from the previous frame. Specifically, during the diffusion process of $(t-1)$-th frame, the predicted noise-free latents $\ddot{y}_{t-1}^n$  at each timestep are stored in the Diffusion Buffer. When reconstructing the $t$-th frame, the corresponding $\ddot{y}_{t-1}^n$ is retrieved directly from the Diffusion Buffer as the predicted noise-free latent $\ddot{y}_{t}^n$ for the current frame, instead of being predicted by the U-Net network. Similar to Independent Diffusion, the mean and variance of the posterior distribution are then computed, followed by posterior sampling to obtain the latent $z_t^{n-1}$ for the next timestep. In contrast to Independent Diffusion, Reuse Diffusion mode bypasses the time-consuming U-Net network and relies solely on a straightforward posterior sampling operation, achieving exceptionally fast diffusion speeds with minimal time consumption.

In summary, the TDIR strategy enhances the inference speed of DiffVC by reducing the number of U-Net computations through the reuse of  diffusion information. It is worth noting that the high perceptual quality achieved by diffusion models is primarily attributed to their iterative sampling process, where fewer sampling steps typically lead to degraded perceptual quality (as shown in Fig. \ref{fig:ds}). However, the TDIR strategy does not reduce the total number of sampling steps. Even during reuse diffusion, it still performs posterior sampling by reusing the predicted noise-free latent from the previous frame, ensuring minimal perceptual quality loss (approximately 1.96\%) while reducing inference time by 47\% (detailed results can be found in the Section \ref{sec:ablation_tdir}). Furthermore, conventional diffusion model acceleration techniques (e.g. DDIM \cite{DBLP:conf/iclr/SongME21}, better start strategy) can be seamlessly integrated with the TDIR strategy for further speed improvements. Notably, the TDIR strategy is also applicable to other diffusion-based video tasks, such as video generation.

\subsection{QP-based Prompting Mechanism} \label{sec:QPP}
Variable bitrate is a fundamental feature of video codecs and is essential for practical applications. In DiffVC, the variable bitrate solution builds on DCVC-DC by introducing quantization parameters ($q_{enc}$ and $q_{dec}$) in the Contextual Encoder/Decoder to scale latent representations. However, the distortion levels vary significantly across different bitrate points, resulting in notable differences in the distributions of latent representations. Employing a single diffusion model to recover latent representations across all bitrate points often yields suboptimal results. A straightforward solution is to train a separate diffusion model for each bitrate point. However, this is impractical due to the large parameter size of the foundation diffusion model, which incurs prohibitively high training costs. Moreover, such a strategy undermines the very concept of variable bitrate encoding. To resolve this, we propose a QP-based Prompting mechanism (QPP) that enables a single diffusion model to support all bitrate points, thereby achieving robust variable bitrate in diffusion-based video compression.

\begin{table*}[!t]
	\caption{Training strategy of our proposed diffusion-based video compression scheme.}
	\renewcommand{\arraystretch}{1.5}
	\centering
	\resizebox{1.0\linewidth}{!}{
		\begin{tabular}{c|c|c|c|c|c|l}
			\toprule[1.5pt]
			Stage  & Tainable Modules & $\lambda$       & Epoch & lr milestones   & lr list      																   & Loss                  	           \\ \hline
			1      & Motion           & 384             & 5     & [0]             & [$10^{-4}$]  																   & $\mathcal{L}_{motion}^{D}$   	   \\ \hline
			2      & Contextual       & 384             & 5     & [0]             & [$10^{-4}$]  																   & $\mathcal{L}_{contextual}^{D}$    \\ \hline
			3      & Motion           & 384             & 8     & [0]             & [$10^{-4}$]   																   & $\mathcal{L}_{motion}^{RD}$  	   \\ \hline
			4      & Contextual       & 384             & 8     & [0]             & [$10^{-4}$]  																   & $\mathcal{L}_{contextual}^{RD}$   \\ \hline
			5      & Motion+Contextual& 384             & 5     & [0,2,3,4]       & [$10^{-4}$,$5\times 10^{-5}$,$10^{-5}$,$5\times 10^{-6}$]                      & $\mathcal{L}_{compress}^{RD}$     \\ \hline
			6      & Motion+Contextual& 16,48,128,384   & 20    & [0,8,12,16,18]  & [$10^{-4}$,$5\times 10^{-5}$,$10^{-5}$,$5\times 10^{-6}$,$10^{-6}$]            & $\mathcal{L}_{compress}^{RDP}$    \\ \hline
			7      & Motion+Contextual& 16,48,128,384   & 10    & [0,4,7,9]       & [$5\times 10^{-5}$,$10^{-5}$,$5\times 10^{-6}$,$10^{-6}$]  					   & $\mathcal{L}_{compress}^{RDPcas}$ \\ \hline
			8      & Diffusion        & 16,48,128,384   & 50    & [0,30,38,45,48] & [$5\times 10^{-5}$,$2.5\times 10^{-5}$,$10^{-5}$,$5\times 10^{-6}$,$10^{-6}$]  & $\mathcal{L}_{diffusion}$    	   \\
			\bottomrule[1.5pt]
		\end{tabular}
	}
	\label{tab:training_stategy}
\end{table*}

Stable Diffusion, originally developed for text-to-image generation, is trained on a large dataset of text-image pairs, giving it a robust understanding of diverse prompts. It is intuitive to leverage Stable Diffusion's built-in semantic understanding to make it explicitly aware of bitrate variations. As shown in Fig. \ref{fig:QPP}, QPP adopts a simple yet effective approach to achieve this. Specifically, $q_{enc}$ and $q_{dec}$ are quantization parameters in the Contextual Encoder/Decoder that control variable bitrates. Their ratio, averaged along the channel dimension, provides a value that characterizes bitrate changes (illustrated by the curve in the top-right corner of Fig. \ref{fig:QPP}, which demonstrates the proportional relationship between this ratio and the bitrate). This ratio is then treated as a prompt and encoded into tokens by the pre-trained CLIP Text Encoder \cite{DBLP:conf/icml/RadfordKHRGASAM21}:
\begin{equation}
	\label{eq:qpp}
	tokens = CLIP(CAP(\frac{q_{enc}}{q_{dec}}))
\end{equation}
where $CLIP$ denotes the pretrained CLIP Text Encoder and $CAP$ represents channel-wise average pooling. These tokens are used to modulate the intermediate features of the U-Net through cross-attention layers, enabling the diffusion model to explicitly perceive distortion variations across different bitrates.

While Stable Diffusion is capable of interpreting semantic information, it does not inherently associate the QP-based prompt with the distortion variations in the latent representation. To address this limitation, we employ targeted training by fine-tuning the diffusion model using a mixed bitrate approach. This fine-tuning enables the diffusion model to establish a robust correspondence between the QP-based prompt and the distortion characteristics of the latent representation (details are provided in the next section).

\subsection{Training Strategy} \label{sec:training_stategy}

To ensure comprehensive training of each module in DiffVC, we adopt a multi-stage training strategy. The specifics of this strategy are outlined in Table \ref{tab:training_stategy}. As described in Section \ref{sec:framework}, DiffVC comprises three main components: Motion, Contextual, and Diffusion modules. The training process is divided into eight distinct stages, with each stage utilizing tailored loss functions to target specific modules. The details of each stage are as follows:

\begin{figure*}[!t]
	\centering
	\includegraphics[width=1.0\textwidth]{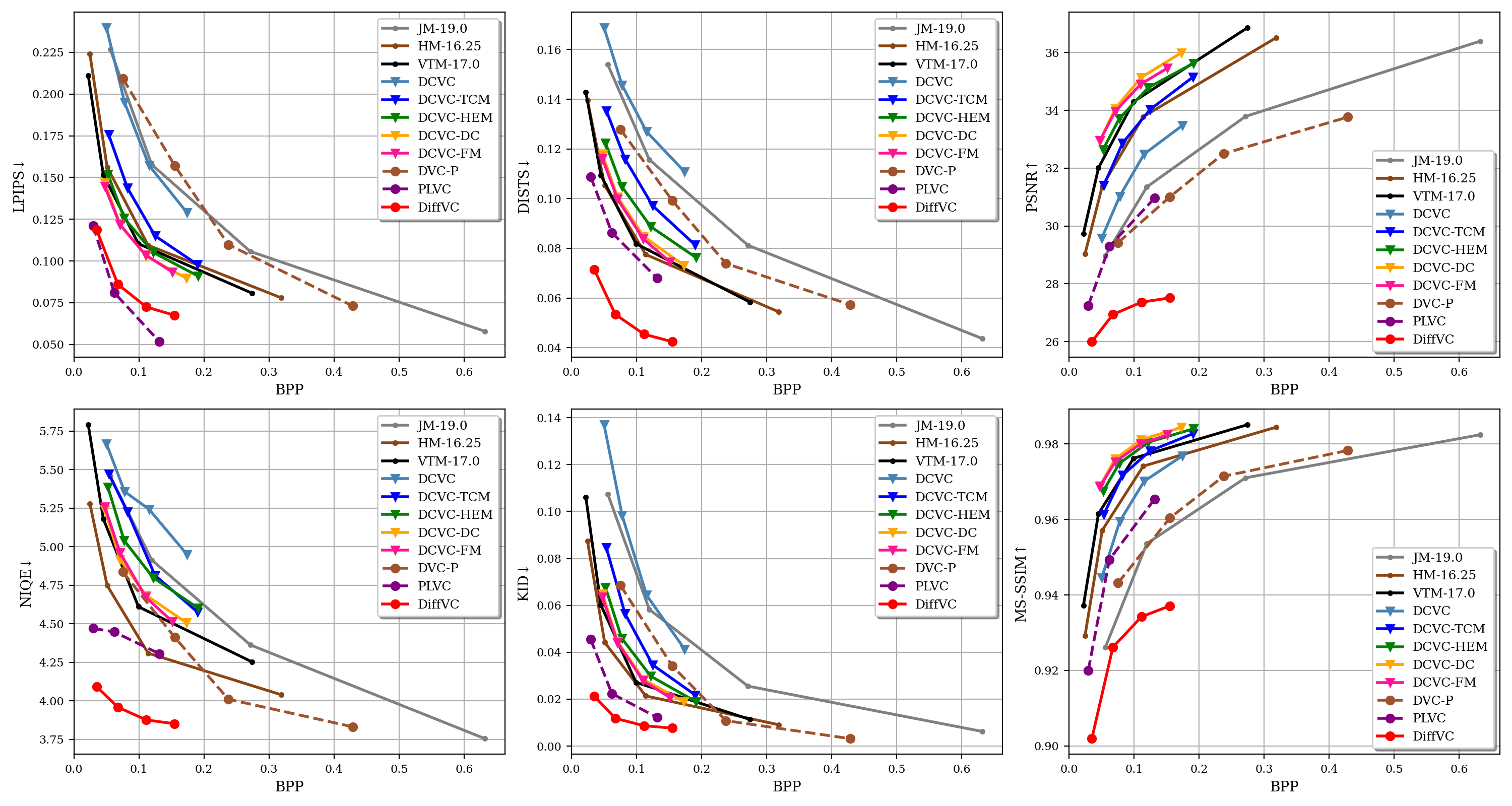}
	\vspace{-20pt}
	\caption{The rate-perception/distortion curves of our proposed DiffVC and other video compression methods on the HEVC dataset. Solid lines with dots represent traditional codecs, solid lines with triangles denote distortion-oriented neural video compression methods, dashed lines with circles indicate GAN-based neural video compression methods, and solid lines with circles correspond to Diffusion-based neural video compression methods.}
	\vspace{-10pt}
	\label{fig:HEVC_RD_Curve}
\end{figure*}

\begin{itemize}
	\item{\textbf{Stage 1:} The Motion Modules are trained at the highest bitrate. As described in Eq. \ref{eq:loss1}, $\mathcal{L}_{motion}^{D}$ calculates the distortion between the warped frame $\check{x}_t$ and the input frame $x_t$, enabling the Motion Modules to reconstruct high-fidelity motion vector.}
	\begin{equation}
		\label{eq:loss1}
		\mathcal{L}_{motion}^{D} = w_t \cdot \lambda \cdot D(x_t, \check{x}_t)
	\end{equation}
	\item{\textbf{Stage 2:} The Contextual Modules are trained at the highest bitrate. As described in Eq. \ref{eq:loss2}, $\mathcal{L}_{contextual}^{D}$ calculates the distortion between the reconstructed frame $\hat{x}_t$ and $x_t$, enabling the Contextual Modules learn to generate high-fidelity results.}
	\begin{equation}
		\label{eq:loss2}
		\mathcal{L}_{contextual}^{D} = w_t \cdot \lambda \cdot D(x_t, \hat{x}_t)
	\end{equation}
	\item{\textbf{Stage 3:} The Motion Modules are further trained at the highest bitrate using the rate-distortion loss $\mathcal{L}_{motion}^{RD}$, which incorporates both the bitrate of motion vector $m_t$ and the distortion loss $\mathcal{L}_{motion}^{D}$.}
	\begin{equation}
		\label{eq:loss3}
		\mathcal{L}_{motion}^{RD} = R(m_t)+ w_t \cdot \lambda \cdot D(x_t, \check{x}_t)
	\end{equation}
	\item{\textbf{Stage 4:} The Contextual Modules are trained at the highest bitrate using the rate-distortion loss $\mathcal{L}_{contextual}^{RD}$, which accounts for both the bitrate of latent representation $y_t$ and the distortion loss $\mathcal{L}_{contextual}^{D}$.}
	\begin{equation}
		\label{eq:loss4}
		\mathcal{L}_{contextual}^{RD} = R(y_t)+ w_t \cdot \lambda \cdot D(x_t, \hat{x}_t)
	\end{equation}
	\item{\textbf{Stage 5:} Both the Motion and Contextual Modules are jointly trained at the highest bitrate. The rate-distortion loss $\mathcal{L}_{compress}^{RD}$, defined in Eq. \ref{eq:loss5}, extends $\mathcal{L}_{contextual}^{RD}$ by incorporating the bitrate of $m_t$, enabling joint optimization of the entire compression network.}
	\begin{equation}
		\label{eq:loss5}
		\mathcal{L}_{compress}^{RD} = R(m_t) + R(y_t) + w_t \cdot \lambda \cdot D(x_t, \hat{x}_t)
	\end{equation}
	\item{\textbf{Stage 6:} The Motion and Contextual Modules are trained across all bitrate levels. The rate-distortion-perception loss $\mathcal{L}_{compress}^{RDP}$, defined in Eq. \ref{eq:loss6}, extends $\mathcal{L}_{compress}^{RD}$ by incorporating a VGG-based loss term to improve the perceptual quality of the reconstructed results.}
	\begin{align}
		\label{eq:loss6}
		\mathcal{L}_{compress}^{RDP} =& R(m_t) + R(y_t) + \notag \\
		&w_t \cdot \lambda \cdot (D(x_t, \hat{x}_t) + w_p \cdot VGG(x_t, \hat{x}_t))
	\end{align}
	\item{\textbf{Stage 7:} The trainable modules are the same as Stage 6. Following \cite{DBLP:journals/tmm/ShengLLLLL23}, this stage employs a cascade training strategy. $\mathcal{L}_{compress}^{RDPcas}$ calculates the average Rate-Distortion-Perception (RDP) loss across $T$ frames, effectively mitigating the accumulation of errors.}
	\begin{align}
		\label{eq:loss7}
		\mathcal{L}_{compress}^{RDPcas} =& \frac{1}{T} \sum_t^T (R(m_t) + R(y_t) + w_t \cdot \lambda \cdot (D(x_t, \hat{x}_t) + \notag \\
		&w_p \cdot VGG(x_t, \hat{x}_t)))
	\end{align}
	\item{\textbf{Stage 8:} Train the Diffusion Modules across all bitrate levels. During this stage, the ControlNet is fully trainable, while only the attention layers of the U-Net (approximately 15\% of the total weights) are fine-tuned. As described in Eq. \ref{eq:loss8}, $\mathcal{L}_{diffusion}$ employs the commonly used MSE loss for diffusion models. Additionally, to enable the diffusion model to interpret QP-based prompts, the corresponding prompt for each bitrate level is provided during each training step.}
	\begin{equation}
		\label{eq:loss8}
		\mathcal{L}_{diffusion} = MSE(\epsilon^n, \epsilon_{\theta}^n)
	\end{equation}
\end{itemize}

We use $\lambda$ to balance the trade-off between bitrate and quality (distortion and perception). Additionally, following \cite{DBLP:conf/cvpr/LiLL23}, we introduce a periodically varying weight, $w_t$, to mitigate error propagation. And $w_p$ represents the weight of the perceptual loss term. It is important to note that throughout the training process, both the encoder $\mathcal{E}$ and the decoder $\mathcal{D}$ remain fixed and are not involved in training.

\section{Experiments}  \label{sec:experiments}
\begin{figure*}[!t]
	\centering
	\includegraphics[width=1.0\textwidth]{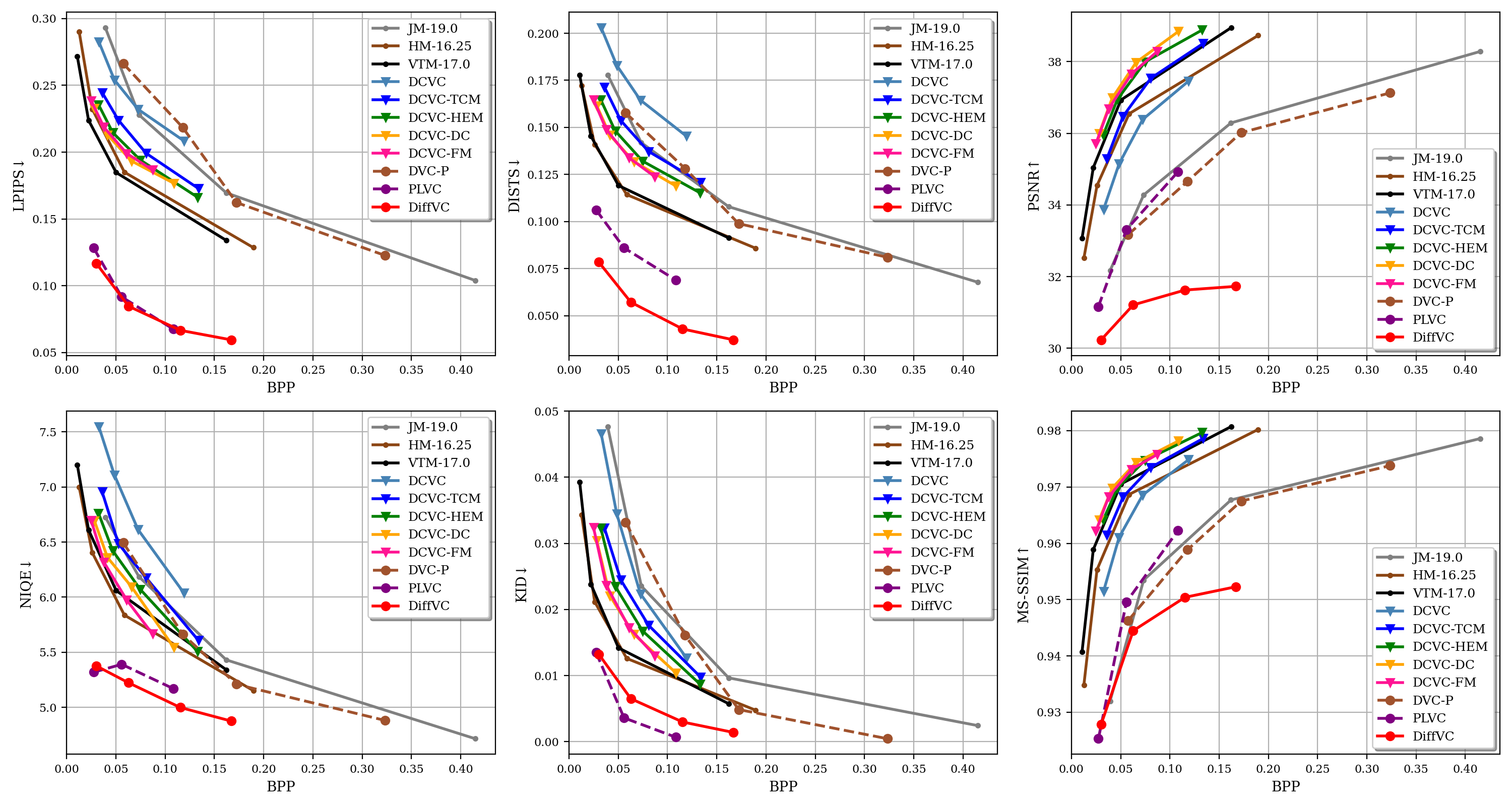}
	\vspace{-20pt}
	\caption{The rate-perception/distortion curves of our proposed DiffVC and other video compression methods on the MCL-JCV dataset. Solid lines with dots represent traditional codecs, solid lines with triangles denote distortion-oriented neural video compression methods, dashed lines with circles indicate GAN-based neural video compression methods, and solid lines with circles correspond to Diffusion-based neural video compression methods.}
	\vspace{-10pt}
	\label{fig:MCL-JCV_RD_Curve}
\end{figure*}
\begin{figure*}[!t]
	\centering
	\includegraphics[width=1.0\textwidth]{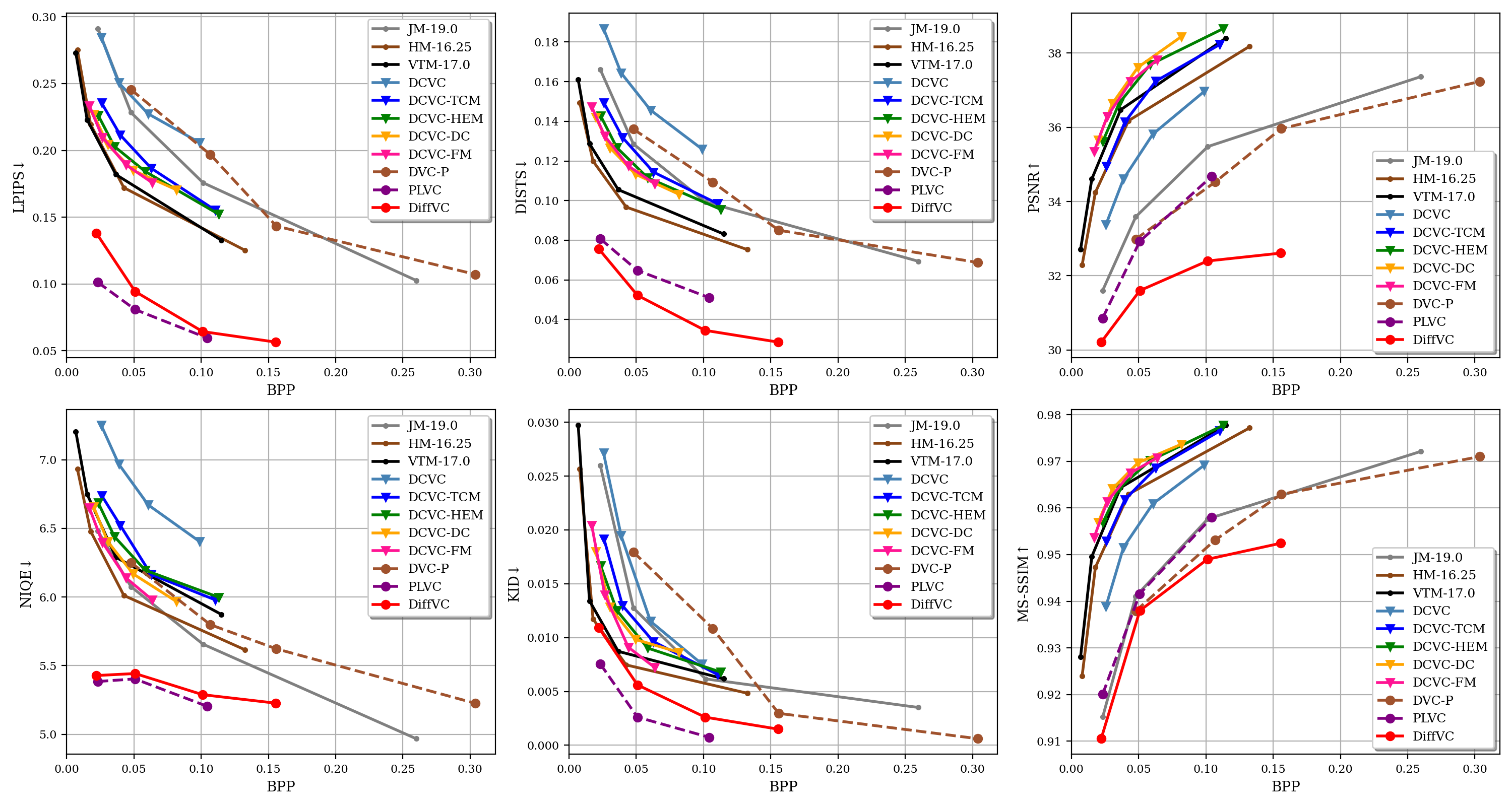}
	\vspace{-20pt}
	\caption{The rate-perception/distortion curves of our proposed DiffVC and other video compression methods on the UVG dataset. Solid lines with dots represent traditional codecs, solid lines with triangles denote distortion-oriented neural video compression methods, dashed lines with circles indicate GAN-based neural video compression methods, and solid lines with circles correspond to Diffusion-based neural video compression methods.}
	\vspace{-10pt}
	\label{fig:UVG_RD_Curve}
\end{figure*}

\begin{table*}[t]
	\caption{BD-rate$\downarrow$ (\%) / BD-metric$\uparrow$ for different methods on HEVC, MCL-JCV and UVG dataset. The anchor is VTM-17.0.}
	\label{tab:result}
	\begin{center}
		\scalebox{0.85}{
			\begin{threeparttable}
				\begin{tabular}{l|l|llll|ll}
					\toprule
					\multirow{2}{*}{Dataset} &\multirow{2}{*}{Methods} &\multicolumn{4}{c|}{Perception} &\multicolumn{2}{c}{Distortion} \\ \cline{3-8}
					&        &DISTS   &LPIPS     &KID    &NIQE   &PSNR    &MS-SSIM \\ \midrule
					\multirow{11}{*}{HEVC} 			
					&JM-19.0  & 176.3 / -0.0384 & 165.0 / -0.0547 & 150.9 / -0.0354 & 76.0 / -0.3532 & 219.8 / -3.4148 & 259.8 / -0.0254 \\
					&HM-16.25 & 0.7 / -0.0002 & 17.6 / -0.0088 & -17.0 / 0.0056 & -39.5 / 0.2806 & 37.8 / -0.9235 & 33.0 / -0.0057 \\
					&VTM-17.0 & 0.0 / 0.0000 & 0.0 / 0.0000 & 0.0 / 0.0000 & 0.0 / 0.0000 & 0.0 / 0.0000 & 0.0 / 0.0000 \\
					&DCVC 	  & 283.4 / -0.0532 & 186.0 / -0.0647 & 191.3 / -0.0531 & 140.6 / -0.6306 & 129.6 / -2.4412 & 83.2 / -0.0112 \\
					&DCVC-TCM & 105.8 / -0.0249 & 53.1 / -0.0201 & 67.9 / -0.0188 & 81.8 / -0.3897 & 38.7 / -0.9163 & 10.2 / -0.0015 \\ 
					&DCVC-HEM & 53.9 / -0.0140 & 9.2 / -0.0040 & 31.0 / -0.0091 & 54.8 / -0.2843 & -0.3 / -0.0097 & -15.3 / 0.0019 \\
					&DCVC-DC  & 25.2 / -0.0073 & -7.0 / 0.0029 & 15.4 / -0.0049 & 19.2 / -0.1135 & \textcolor{red}{\textbf{-19.8 / 0.5900}} & \textcolor{red}{\textbf{-28.5 / 0.0042}} \\
					&DCVC-FM  & 21.0 / -0.0063 & -7.3 / 0.0032 & 13.5 / -0.0045 & 21.0 / -0.1224 & \textcolor{blue}{\textbf{-16.4 / 0.4788}} & \textcolor{blue}{\textbf{-23.3 / 0.0036}} \\
					&DVC-P    & 122.9 / -0.0278 & 195.2 / -0.0609 & 75.3 / -0.0167 & 1.6 / 0.0385 & 359.0 / -4.2982 & 241.0 / -0.0211 \\
					&PLVC 	  & \textcolor{blue}{\textbf{-27.1 / 0.0110}} & \textcolor{red}{\textbf{-66.9 / 0.0518}} & \textcolor{blue}{\textbf{-47.3 / 0.0216}} & \textcolor{blue}{\textbf{-48.2 / 0.5238}} & 262.4 / -3.7475 & 120.0 / -0.0209 \\
					&DiffVC   & \textcolor{red}{\textbf{-79.0 / 0.0391}} & \textcolor{blue}{\textbf{-64.8 / 0.0401}} & \textcolor{red}{\textbf{-75.8 / 0.0281}} & \textcolor{red}{\textbf{N/A / 0.8849}} & N/A / -6.4813 & N/A / -0.0453 \\ \midrule
					\multirow{11}{*}{MCL-JCV} 	
					&JM-19.0  & 167.5 / -0.0322 & 188.8 / -0.0604 & 190.5 / -0.0134 & 59.8 / -0.3221 & 333.2 / -3.3522 & 315.7 / -0.0215   \\
					&HM-16.25 & 0.6 / -0.0001 & 23.7 / -0.0118 & -2.6 / 0.0003 & -14.2 / 0.1046 & 40.2 / -0.7545 & 37.5 / -0.0049 \\
					&VTM-17.0 & 0.0 / 0.0000 & 0.0 / 0.0000 & 0.0 / 0.0000 & 0.0 / 0.0000 & 0.0 / 0.0000 & 0.0 / 0.0000  \\
					&DCVC     & 419.1 / -0.0584 & 269.1 / -0.0665 & 211.9 / -0.0152 & 221.5 / -0.8732 & 99.8 / -1.4660 & 83.8 / -0.0075 \\
					&DCVC-TCM & 190.5 / -0.0328 & 124.3 / -0.0378 & 125.7 / -0.0087 & 88.7 / -0.4180 & 24.9 / -0.4232 & 24.5 / -0.0024  \\ 
					&DCVC-HEM & 128.7 / -0.0249 & 79.3 / -0.0270 & 93.4 / -0.0069 & 47.6 / -0.2600 & -10.9 / 0.2178 & -1.3 / -0.0000 \\
					&DCVC-DC  & 95.8 / -0.0208 & 56.6 / -0.0213 & 64.0 / -0.0053 & 27.4 / -0.1681 & \textcolor{red}{\textbf{-22.3 / 0.4831}} & \textcolor{red}{\textbf{-10.4 / 0.0011}} \\
					&DCVC-FM  & 90.2 / -0.0210 & 58.5 / -0.0224 & 67.8 / -0.0061 & 9.5 / -0.0607 & \textcolor{blue}{\textbf{-16.1 / 0.3549}} & \textcolor{blue}{\textbf{-7.0 / 0.0008}} \\
					&DVC-P    & 160.8 / -0.0334 & 236.5 / -0.0753 & 186.6 / -0.0117 & 48.6 / -0.2384 & 465.4 / -3.8693 & 388.0 / -0.0214 \\
					&PLVC     & \textcolor{blue}{\textbf{-67.6 / 0.0300}} & \textcolor{blue}{\textbf{N/A / 0.0857}} & \textcolor{red}{\textbf{-58.3 / 0.0082}} & \textcolor{blue}{\textbf{-- / 0.6759}} & 384.6 / -3.9255 & 296.0 / -0.0247 \\
					&DiffVC   & \textcolor{red}{\textbf{N/A / 0.0556}} & \textcolor{red}{\textbf{N/A / 0.0868}} & \textcolor{blue}{\textbf{-52.5 / 0.0054}} & \textcolor{red}{\textbf{-78.8 / 0.6902}} & N/A / -6.3123 & 509.2 / -0.0296 \\ \midrule
					\multirow{11}{*}{UVG} 		
					&JM-19.0  & 171.1 / -0.0285 & 182.7 / -0.0582 & 166.0 / -0.0053 & -19.6 / 0.1257 & 363.0 / -3.3050 & 339.1 / -0.0274  \\
					&HM-16.25 & -16.1 / 0.0048 & 3.6 / -0.0014 & -5.7 / 0.0005 & -33.5 / 0.1992 & 36.0 / -0.6329 & 29.2 / -0.0046 \\
					&VTM-17.0 & 0.0 / 0.0000 & 0.0 / 0.0000 & 0.0 / 0.0000 & 0.0 / 0.0000 & 0.0 / 0.0000 & 0.0 / 0.0000  \\
					&DCVC     & 527.7 / -0.0552 & 326.3 / -0.0710 & 248.9 / -0.0076 & 263.9 / -0.6372 & 143.1 / -1.7526 & 125.2 / -0.0121 \\
					&DCVC-TCM & 165.7 / -0.0237 & 94.8 / -0.0300 & 126.9 / -0.0036 & 50.5 / -0.1728 & 22.4 / -0.3524 & 27.2 / -0.0032 \\ 
					&DCVC-HEM & 112.3 / -0.0178 & 62.1 / -0.0215 & 90.5 / -0.0026 & 32.6 / -0.1210 & -14.0 / 0.2809 & 0.4 / -0.0002 \\
					&DCVC-DC  & 89.4 / -0.0160 & 43.5 / -0.0168 & 85.3 / -0.0031 & 3.8 / -0.0182 & \textcolor{red}{\textbf{-25.8 / 0.5460}} & \textcolor{red}{\textbf{-11.6 / 0.0015}} \\
					&DCVC-FM  & 95.1 / -0.0181 & 38.1 / -0.0151 & 65.9 / -0.0030 & -12.3 / 0.0578 & \textcolor{blue}{\textbf{-20.4 / 0.4366}} & \textcolor{blue}{\textbf{-8.1 / 0.0011}} \\
					&DVC-P    & 199.4 / -0.0325 & 226.3 / -0.0728 & 306.0 / -0.0080 & -0.8 / 0.0186 & 528.6 / -4.0096 & 434.7 / -0.0274 \\
					&PLVC     & \textcolor{blue}{\textbf{N/A / 0.0336}} & \textcolor{red}{\textbf{N/A / 0.0873}} & \textcolor{red}{\textbf{-66.4 / 0.0045}} & \textcolor{red}{\textbf{N/A / 0.8133}} & 568.5 / -4.2150 & 374.9 / -0.0284 \\
					&DiffVC   & \textcolor{red}{\textbf{N/A / 0.0464}} & \textcolor{blue}{\textbf{-78.6 / 0.0722}} & \textcolor{blue}{\textbf{-27.4 / 0.0020}} & \textcolor{blue}{\textbf{N/A / 0.7705}} & N/A / -5.5030 & 486.3 / -0.0330 \\
					\bottomrule
				\end{tabular}
				\begin{tablenotes}
					\item[*] \textbf{\textcolor{red}{Red}} and \textbf{\textcolor{blue}{Blue}} indicate the best and the second-best performance, respectively. 'N/A' indicates that BD-rate cannot be calculated due to the lack of overlap. '--' indicates that BD-rate cannot be calculated because the rate-NIQE curve is not monotonic.
				\end{tablenotes}
			\end{threeparttable}
		}
	\end{center}
	\vspace{-10pt}
\end{table*}

\subsection{Experimental Setup}
\paragraph{Datasets} For training, we use the widely adopted Vimeo-90k dataset \cite{DBLP:journals/ijcv/XueCWWF19}, which is commonly used for video compression research. This dataset comprises 89,800 video clips containing diverse real-world motions, with each clip consisting of seven consecutive frames. To comprehensively evaluate the performance of our proposed video compression framework, we employ HEVC \cite{DBLP:journals/tcsv/SullivanOHW12}, UVG \cite{DBLP:conf/mmsys/MercatVV20} and MCL-JCV \cite{DBLP:conf/icip/WangGHLJSWKAK16} as test datasets. These datasets include video resolutions of 416$\times$240, 832$\times$480, 1280$\times$720, and 1920$\times$1080, providing a diverse range of scenes for comprehensive evaluation.

\paragraph{Training Settings} As outlined in Section \ref{sec:training_stategy}, we employ a multi-stage training strateg. To support variable bitrate with a single model, we define four $\lambda$ values (16, 48, 128, 384) to balance bitrate and reconstruction quality. Following \cite{DBLP:conf/cvpr/LiLL23}, the periodically varying weights $w_t$ for four consecutive frames are set to (0.5, 1.2, 0.5, 0.9). The perceptual loss weight $w_p$ is set to 0.025. Training is performed with a batch size of 8, and, as is common in video compression methods, sequences are randomly cropped to a resolution of 256$\times$256. The Adam optimizer is used with $\beta_1=0.9$ and $\beta_2=0.999$. The learning rate is initialized at $1\times10^{-4}$ and the specific learning rate decay strategy during training is detailed in columns 5 and 6 of Table \ref{tab:training_stategy}. All experiments are implemented in PyTorch and conducted using NVIDIA GTX 3090 GPUs.

\paragraph{Test Settings} Following \cite{DBLP:journals/tmm/ShengLLLLL23,DBLP:conf/mm/Li0022,DBLP:conf/cvpr/LiLL23}, we evaluate the first 96 frames of each video sequence, with the intra-period set to 32. The low-delay encoding configuration is used, and HIFIC \cite{DBLP:conf/nips/MentzerTTA20} is employed to encode I frames. For the diffusion model, the total diffusion steps ($DS$) is set to 50, while the independent diffusion steps ($D$) in TDIR is set to 25. All evaluations are conducted in the RGB color space.

\paragraph{Compared Methods} We conduct a comprehensive comparison of the proposed DiffVC with various available video compression methods, including traditional reference software: JM-19.0 \cite{JM}, HM-16.25 \cite{HM} and VTM-17.0 \cite{VTM}; distortion-oriented neural video compression methods: DCVC \cite{DBLP:conf/nips/LiLL21}, DCVC-TCM \cite{DBLP:journals/tmm/ShengLLLLL23}, DCVC-HEM \cite{DBLP:conf/mm/Li0022}, DCVC-DC \cite{DBLP:conf/cvpr/LiLL23} and DCVC-FM \cite{DBLP:conf/cvpr/LiLL24}; GAN-based perceptual neural video compression methods: DVC-P \cite{DBLP:conf/vcip/ZhangMHBWY21} and PLVC \cite{DBLP:conf/ijcai/YangTG22}; and diffusion-based perceptual neural video compression method: EVC-PDM \cite{DBLP:journals/corr/abs-2402-08934}. For JM, HM, and VTM, we use the \textit{encoder\_baseline}, \textit{encoder\_lowdelay\_main\_rext} and \textit{encoder\_lowdelay\_vtm} configurations, QP values for the four bitrate points are 22, 27, 32, and 37. For DVC-P, we retrain the model according to its training procedure, as its pre-trained model is not available, and follow the original intra-period of 10. For PLVC, we also set the intra-period to 9 as specified in the original paper. For EVC-PDM, which only supports video inputs with a resolution of 128$\times$128, we align its test conditions as described in Section \ref{sec:complementary_comparison} and compare it with our method.

\paragraph{Evaluation Metrics} To quantitatively evaluate performance, we use several established metrics to assess the quality of the reconstructed videos. For distortion metrics, we adopt Peak Signal-to-Noise Ratio (PSNR) and Multi-Scale Structural Similarity Index Measure (MS-SSIM) \cite{wang2003multiscale} to evaluate the fidelity of the reconstructed results. For perceptual metrics, we use the reference Learned Perceptual Image Patch Similarity (LPIPS) \cite{DBLP:conf/cvpr/ZhangIESW18}, Deep Image Structure and Texture Similarity (DISTS) \cite{DBLP:journals/pami/DingMWS22}, Kernel Inception Distance (KID) \cite{DBLP:conf/iclr/BinkowskiSAG18}, and the non-reference Natural Image Quality Evaluator (NIQE) \cite{DBLP:journals/spl/MittalSB13} to comprehensively measure the perceptual quality of the reconstructed outputs. Finally, we use Bit Per Pixel (BPP) to measure the bits cost for encoding one pixel in each frame.

\subsection{Experimental Results}
\begin{figure*}[!t]
	\centering
	\includegraphics[width=1.0\textwidth]{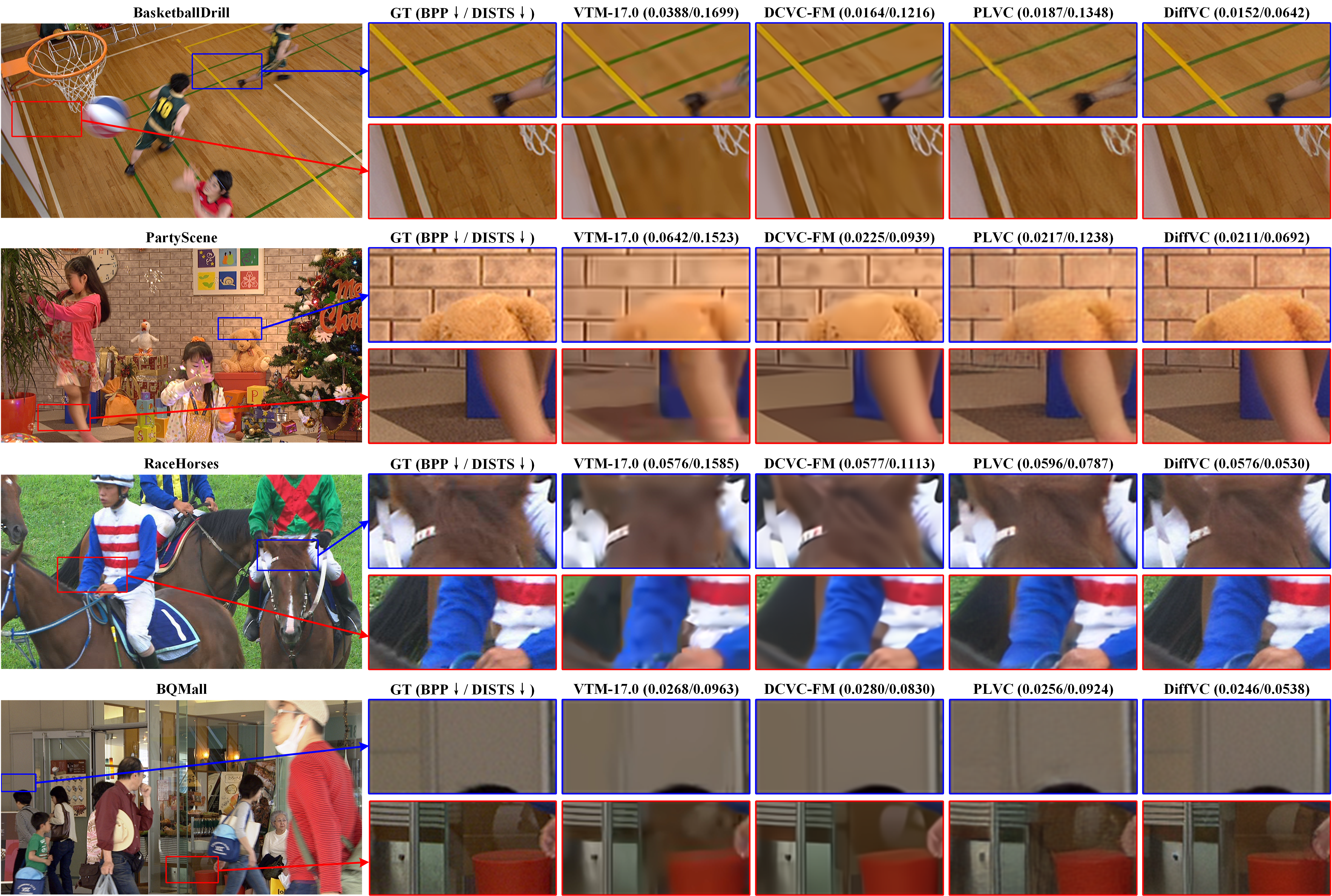}
	\caption{Visual results of traditional codec VTM-17.0, distortion-oriented codec DCVC-FM, GAN-based codec PLVC and diffusion-based DiffVC (ours).}
	\label{fig:visual}
\end{figure*}

\paragraph{Perception Metric Evaluation} Fig. \ref{fig:HEVC_RD_Curve}, \ref{fig:MCL-JCV_RD_Curve} and \ref{fig:UVG_RD_Curve} present the rate-perception/distortion curves for various video compression methods on the HEVC, MCL-JCV, and UVG datasets, respectively. In each figure, the first and second columns evaluate reconstruction quality using perception metrics, while the last column assesses it using distortion metrics. Table \ref{tab:result} further reports the BD-rate and BD-metric for all metrics, with VTM-17.0 as the anchor. Since the perception metrics used in this paper follow a lower-is-better convention, their values are negated during the calculation of BD-rate and BD-metric for clarity. Consequently, smaller BD-rate values and larger BD-metric values indicate better performance. Experimental results demonstrate that the proposed DiffVC consistently outperforms all traditional and distortion-oriented neural video compression methods in perception metrics. Compared to GAN-based methods, DiffVC achieves superior results across most perception metrics, particularly excelling in the DISTS metric. For example, on the HEVC dataset, DiffVC significantly outperforms PLVC in the DISTS, KID and NIQE metrics, while ranking second in LPIPS with a minimal gap of only 2.1\%. Overall, DiffVC exhibits exceptional performance in perception metrics.

\paragraph{Visual Results} This section presents the visual quality of reconstructed results from different video compression methods. Four representative methods from distinct video compression categories are selected for a comprehensive comparison: the traditional codec VTM-17.0, the distortion-oriented neural codec DCVC-FM, the GAN-based neural codec PLVC, and our proposed diffusion-based neural codec DiffVC. Fig. \ref{fig:visual} shows the reconstruction results from these methods, using four video sequences from HEVC Class C: \textit{BasketballDrill}, \textit{PartyScene}, \textit{RaceHorses} and \textit{BQMall}. A detailed comparison focusing on the floor in \textit{BasketballDrill}, the doll's head and carpet in \textit{PartyScene}, the horse's tail and mane in \textit{RaceHorses} and the textures on the wall in \textit{BQMall} reveals that our proposed DiffVC produces a more detailed and visually appealing reconstruction at a lower bitrate, outperforming the other methods.

\paragraph{Complementary Comparison}\label{sec:complementary_comparison}
Similar to DiffVC, EVC-PDM is a diffusion-based perceptual video compression method that combines image compression with diffusion models to achieve high perceptual quality in extreme scenarios. Since EVC-PDM only supports inputs with a resolution of 128$\times$128, this section aligns the testing conditions to compare DiffVC with EVC-PDM. We selected the Class D subset from the HEVC dataset, which contains videos with the smallest resolution. Following \cite{DBLP:journals/corr/abs-2402-08934}, we center-cropped and downsampled the original frames to 128×128 resolution to serve as both input and ground truth, denoted as \textit{HEVC\_ClassD\_128}. Table \ref{tab:evc_pdm_bdrate} summarizes the BD-rate for DiffVC, with EVC-PDM as the anchor. Notably, SSIM is used instead of MS-SSIM, as the latter cannot be computed at a resolution of 128×128 when the gaussian kernel size is set to the default value of 11. Across both perceptual and distortion metrics, DiffVC consistently outperforms EVC-PDM.
\begin{table}[t]
	\caption{BD-rate$\downarrow$ (\%) for DiffVC on the \textit{HEVC\_ClassD\_128} dataset. The anchor is EVC-PDM.}
	\label{tab:evc_pdm_bdrate}
	\begin{center}
		\scalebox{0.9}{
			\begin{threeparttable}
				\begin{tabular}{l|llll|ll}
					\toprule
					\multirow{2}{*}{Methods} &\multicolumn{4}{c|}{Perception} &\multicolumn{2}{c}{Distortion} \\ \cline{2-7}
					&DISTS &LPIPS &KID &NIQE &PSNR &SSIM \\ \midrule
					DiffVC       &-78.2 &-78.4  &-86.6 &N/A &-68.7 &-61.8 \\
					\bottomrule
				\end{tabular}
				\begin{tablenotes}
					\item[*] 'N/A' indicates that BD-rate cannot be calculated due to the lack of overlap.
				\end{tablenotes}
			\end{threeparttable}
		}
	\end{center}
\end{table}

\subsection{Ablation Study}
\begin{table*}[!htbp]
	\centering
	\caption{The ablation study of our proposed DiffVC. We provide the BD-rate$\downarrow$ (\%) for perception metrics and Decoding Time (seconds) for a 480p frame. All results are tested on HEVC Class C. The anchor is DiffVC.}
	\label{tab:ablation}
	\begin{center}
		\scalebox{0.9}{
			\begin{threeparttable}
				\begin{tabular}{c|ccc||ccccc|c}
					\toprule
					Method & Diffusion & TDIR & QPP & DISTS & LPIPS & KID & NIQE & Average & Decoding Time (s)
					\\\midrule
					A      & \XSolidBrush &\XSolidBrush &\XSolidBrush &25.76 &71.40 &216.99 &--     &--    &0.1933 \\
					B      & \Checkmark   &\XSolidBrush &\XSolidBrush &-0.60 &-0.03 &6.34   &-5.72  &0.00  &7.5414 \\
					C      & \Checkmark   &\Checkmark   &\XSolidBrush &3.70  &3.48  &5.84   &7.35   &5.09  &4.0127 \\
					D      & \Checkmark   &\XSolidBrush &\Checkmark   &-3.19 &-2.93 &3.59   &-5.31  &-1.96 &7.5503 \\
					DiffVC & \Checkmark   &\Checkmark   &\Checkmark   &0.00  &0.00  &0.00   &0.00   &0.00  &4.0018 \\\bottomrule
				\end{tabular}
				\begin{tablenotes}
					% Decoding Time表示每个P帧所需的解码时间，I帧的解码时间不是本文的研究重点。
					\item[*] Decoding Time refers to the time required to decode a P frame, as the decoding time for I frame is not the focus of this paper. '--' indicates that BD-rate cannot be calculated because the rate-NIQE curve is not monotonic.
				\end{tablenotes}
			\end{threeparttable}
		}		
	\end{center}
\end{table*}

\begin{figure*}[t]
	\centering
	\includegraphics[width=0.9\textwidth]{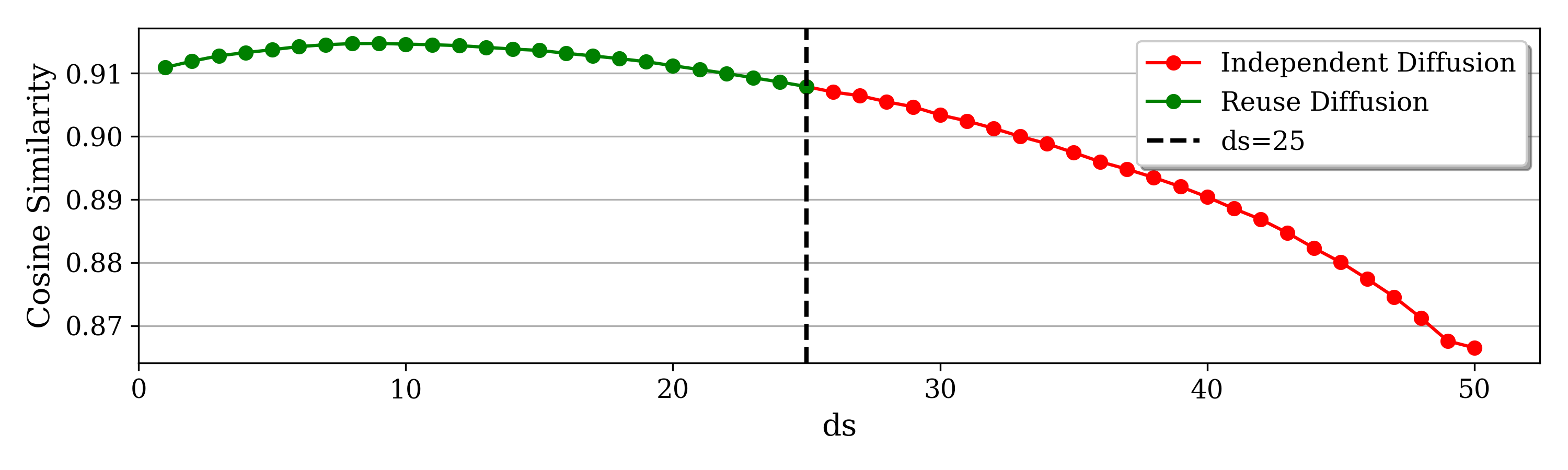}
	\vspace{-20pt}
	\caption{The average cosine similarity of predicted noise-free latent $\ddot{y}_t^n$ between adjacent frames across different diffusion step index ($ds$) when total number of diffusion steps ($DS$) is set to 50.}
	\vspace{-10pt}
	\label{fig:cosine_similarity}
\end{figure*}

\begin{figure*}[t]
	\centering
	\subfloat[]{
		\includegraphics[width=0.48\linewidth]{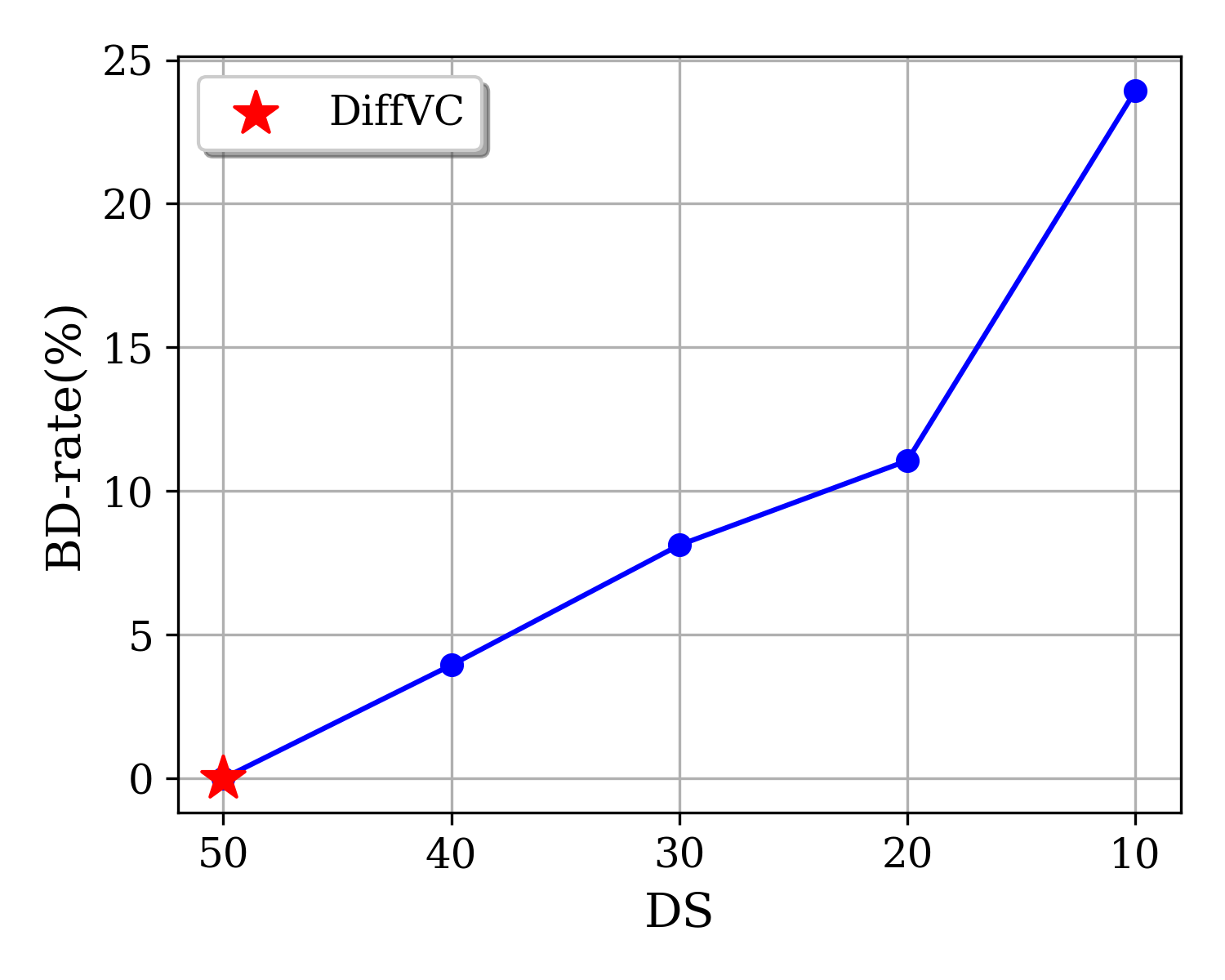}
		\label{fig:ds}
	}\hfill
	\subfloat[]{
		\includegraphics[width=0.48\linewidth]{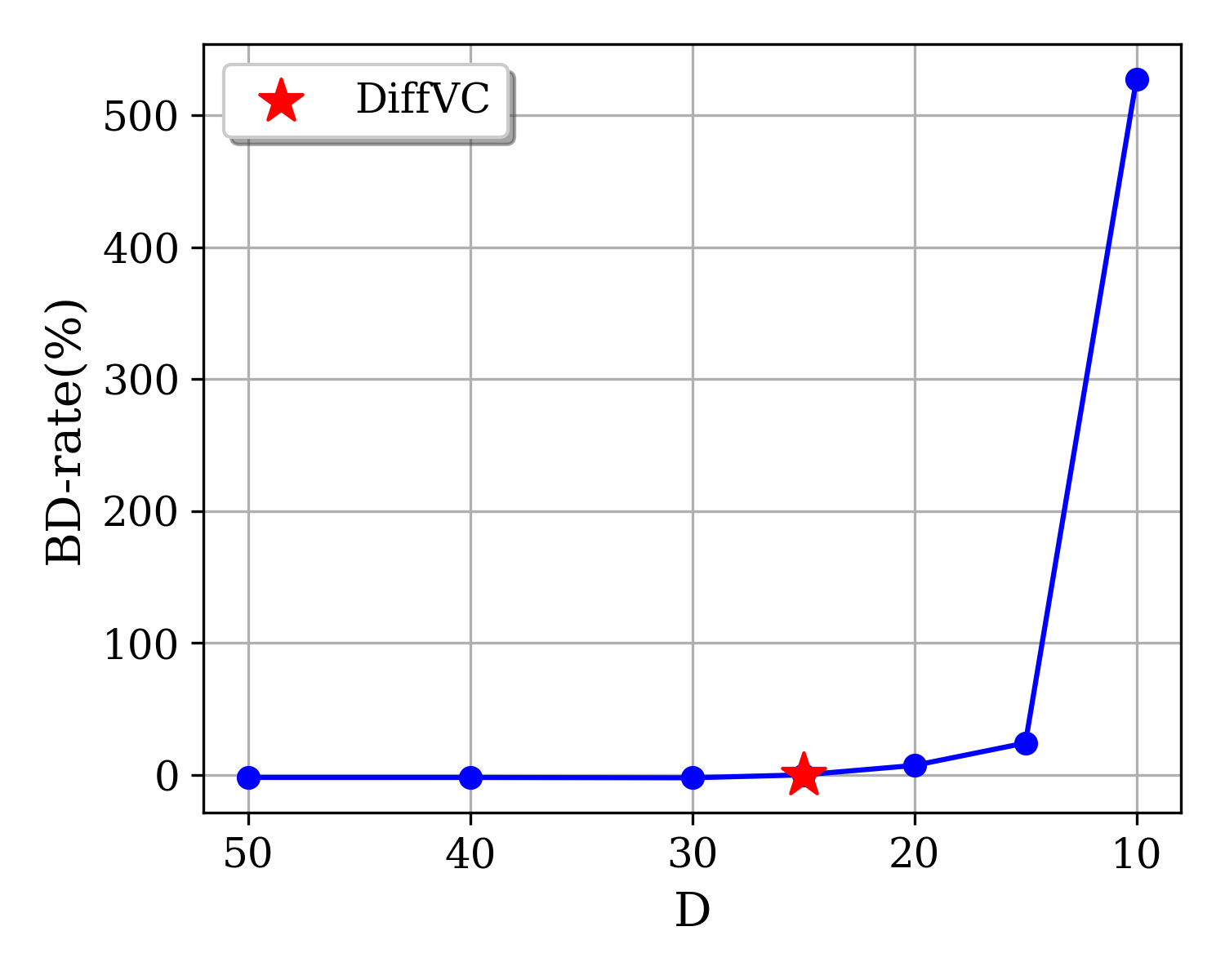}
		\label{fig:d}
	}
	\caption{Influence of $DS$ and $D$. Subfigure (a) illustrates the variation in perceptual BD-rate across different total diffusion steps ($DS$), with DiffVC as the anchor. Subfigure (b) shows the changes in perceptual BD-rate under different independent diffusion steps ($D$) while maintaining $DS=50$, also using DiffVC as the anchor.}
	\label{fig:dicussion_on_tdir}
\end{figure*}

To validate the effectiveness of the proposed modules, we conducted comprehensive ablation experiments. Table \ref{tab:ablation} summarizes the results, evaluating the BD-rate for four perception metrics and the decoding time.
\subsubsection{\textbf{Effectiveness of the Diffusion Model}}
A comparison of Methods A and B in Table \ref{tab:ablation} underscores the critical role of the foundational diffusion model in achieving high-quality reconstruction results. When the diffusion model is not employed (Method A), maintaining the same reconstruction quality requires an increase in bitrate by 26.36\%, 71.43\%, and 210.65\% for the DISTS, LPIPS and KID metrics, respectively. Furthermore, at the same bitrate, the NIQE metric deteriorates by 0.029. This advantage stems from the foundational diffusion model’s pretraining on a vast dataset of high-quality text-image pairs, equipping it with the capability to produce superior results. By leveraging these robust priors, DiffVC achieves exceptional performance in perceptual quality.
\subsubsection{\textbf{Effectiveness of the TDIR strategy}}\label{sec:ablation_tdir}
The primary challenge for diffusion-based video compression methods lies in their significant inference delay. The TDIR strategy significantly accelerates DiffVC's inference speed by reusing diffusion information from the previous frame. A comparison between Method D and DiffVC in Table \ref{tab:ablation} reveals that applying the TDIR strategy reduces inference time by 47\%, with only a minimal perceptual performance loss of 1.96\%. Considering the considerable improvement in inference speed, this slight performance trade-off is deemed acceptable.
\subsubsection{\textbf{Effectiveness of the QPP mechanism}}
In a variable bitrate model, the differences in latent distributions across bitrates make directly employing the diffusion model to recover noisy latent representations suboptimal. To address this, we leverage the foundational diffusion model's inherent semantic understanding by using quantization parameters as prompts to explicitly modulate the intermediate features of the U-Net. This mechanism enables the diffusion model to adaptively reconstruct latent representations with varying levels of distortion across different bitrates, achieving optimal performance. The comparison between Method C and DiffVC in Table \ref{tab:ablation} demonstrates the effectiveness of the QPP mechanism. For perception metrics, the QPP mechanism achieves an average bitrate saving of 5.09\% at the same reconstruction quality. For distortion metrics, it also provides a 5.23\% bitrate saving (6.50\% for PSNR and 3.95\% for MS-SSIM).

\subsubsection{\textbf{Discussion on the TDIR strategy}}

\paragraph{Temporal Correlation of Diffusion Information.}
Fig. \ref{fig:cosine_similarity} demonstrates the high temporal correlation of predicted noise-free latent $\ddot{y}_t^n$ across video frames at different diffusion steps, with cosine similarity consistently exceeding 0.86. This correlation arises from the significant redundancy between video frames, forming the basis for reusing the preticted noise-free latent of previous frames in the TDIR strategy. Moreover, the correlation decreases as the diffusion step index $ds$ increases, justifying the use of reuse-based diffusion in the initial 25 steps to accelerate inference, followed by independent diffusion in later 25 steps to restore frame details. \\

\paragraph{Influence of $\bm{DS}$ and $\bm{D}$.}
Fig. \ref{fig:ds} shows the perceptual performance of the proposed method under different total diffusion steps ($DS$). The results indicate that perceptual performance degrades progressively as $DS$ decreases. Fig. \ref{fig:d} presents the perceptual performance under different independent diffusion steps ($D$). It reveals that perceptual performance remains stable when $D\ge25$, but drops sharply when $D<15$. This is because insufficient independent diffusion steps fail to recover frame-specific details, and excessive reuse of prior frame diffusion information leads to significant error accumulation. In conclusion, DiffVC adopts a configuration of $DS=50$ and $D=25$. \\

\paragraph{Speed of Different Diffusion Modes.}
Table \ref{tab:speed_of_diffuison_mode} presents the speeds of two different diffusion modes in the TDIR strategy. Independent diffusion is over 400 times slower than reuse diffusion. Independent diffusion relies on a large and highly complex U-Net network to estimate noise, resulting in significantly slower inference speeds. In contrast, reuse diffusion bypasses the U-Net network by reusing diffusion information from previous frames and performs only a simple post-sampling operation, enabling much faster inference. This explains why the TDIR strategy achieves efficient diffusion.

\begin{table}[t]
	\caption{Speed of Different Diffusion Modes. Diffusion Time (seconds) refers to the time consumed for a single diffusion step with a 480p video frame as input.}
	\label{tab:speed_of_diffuison_mode}
	\begin{center}
		\scalebox{1.0}{
			\begin{tabular}{l|l}
				\toprule
				Diffusion Mode        & Diffuison Time (s) \\ \midrule
				Reuse Diffusion       & 0.00034 (100\%)    \\
				Independent Diffusion & 0.14154 (41675\%)  \\
				\bottomrule
			\end{tabular}
		}
	\end{center}
\end{table}

\section{Conclusions} \label{sec:conclusions}
In this paper, we propose a diffusion-based perceptual neural video compression framework, named DiffVC. This framework effectively integrates the foundational diffusion model into a conditional coding paradigm, leveraging reconstructed latent representation of the current frame and contextual information mined from previous frames to guide the diffusion model in generating high-perceptual-quality reconstructions. To address the high inference latency inherent in diffusion-based methods, we introduce the TDIR strategy, which significantly accelerates inference by reusing diffusion information from previous frames. Due to the strong temporal correlation between video frames and the robustness of the foundational diffusion model, the performance loss caused by TDIR is minimal. Additionally, to enable the framework to roboustly support variable bitrates, we propose the QPP mechanism. By utilizing quantization parameters as prompts to explicitly modulate features within the U-Net, the diffusion model adapts to distortion variations in latent representations across different bitrates. Extensive experiments conducted on several test datasets demonstrate the effectiveness of our proposed framework.

\appendix

\section*{Appendix}
\section{PARAMETER SETTINGS}
The detailed settings of JM, HM, and VTM are as follows.
\begin{itemize}
	\item \textbf{JM}\\
	lencod.exe -d encoder\_baseline.cfg \\
	-p InputFile=input\_path 
	-p OutputFile=bin\_path \\
	-p SourceWidth=width
	-p SourceHeight=height \\
	-p FrameRate=fps
	-p FramesToBeEncoded=96 \\
	-p IntraPeriod=32 
	-p ProfileIDC=66 \\
	-p YUVFormat=1 
	-p SourceBitDepthLuma=8 \\
	-p SourceBitDepthChroma=8 
	-p QPISlice=28 \\
	-p QPPSlice=28
	\item \textbf{HM}\\
	TAppEncoderStatic \\
	-c encoder\_lowdelay\_main\_rext.cfg \\
	--InputFile=input\_path
	--BitstreamFile=bin\_path \\
	--DecodingRefreshType=2 
	--InputBitDepth=8 \\
	--OutputBitDepth=8 
	--OutputBitDepthC=8 \\
	--InputChromaFormat=444 
	--FrameRate=fps \\
	--FramesToBeEncoded=96 
	--SourceWidth=width \\
	--SourceHeight=height 
	--IntraPeriod=32 \\
	--QP=qp 
	--Level=6.2
	\item \textbf{VTM}\\
	EncoderAppStatic -c encoder\_lowdelay\_vtm.cfg \\
	--InputFile=input\_path 
	--BitstreamFile=bin\_path \\
	--DecodingRefreshType=2 
	--InputBitDepth=8 \\
	--OutputBitDepth=8 
	--OutputBitDepthC=8 \\
	--InputChromaFormat=444 
	--FrameRate=fps \\
	--FramesToBeEncoded=96 
	--SourceWidth=width \\
	--SourceHeight=height 
	--IntraPeriod=32 \\
	--QP=qp 
	--Level=6.2
\end{itemize}

\bibliography{reference}
% that's all folks
\end{document}